\documentclass[11pt]{article}
\usepackage[margin=1in]{geometry}
\usepackage[T1]{fontenc}
\usepackage{lmodern}
\usepackage{authblk}

\usepackage[
  style=authoryear,
  backend=biber,
  giveninits=true,
  uniquename=init
  ]{biblatex}
\AtBeginDocument{\let\cite\parencite}

\usepackage{amsmath,amssymb,amsthm}
\usepackage{booktabs}
\usepackage{graphicx}
\usepackage{subcaption}
\usepackage{longtable}
\usepackage{tabularx}
\usepackage{multirow}
\usepackage{makecell}
\usepackage{siunitx}
\usepackage[table,dvipsnames]{xcolor}
\usepackage{pdflscape}
\usepackage{tikz}
\usetikzlibrary{arrows.meta,positioning,calc,decorations.pathreplacing,shapes.geometric,fit}
\usepackage{algpseudocode}
\usepackage{hyperref}
\usepackage{cleveref}
\usepackage{stfloats}
\usepackage{float}

\title{A Unified Taxonomy and Evaluation Framework for Distributional
Shifts in Reinforcement Learning}

\author[1,2,3]{Ardianto Wibowo\thanks{Corresponding author:
\texttt{ardianto.wibowo@imt-atlantique.fr}}}
\author[4,3]{Paulo E. Santos}
\author[1,3]{Amer Baghdadi}
\author[2,3]{Matthew Stephenson}
\author[2,3]{Karl Sammut}
\author[5,3]{Jean-Philippe Diguet}

\affil[1]{IMT Atlantique, Brest, France}
\affil[2]{Flinders University, Adelaide, Australia}
\affil[3]{IRL Crossing, Adelaide, Australia}
\affil[4]{Priori Analytica, Adelaide, Australia}
\affil[5]{CNRS, France}
\date{}

\begin{document}
\maketitle

\begin{abstract}

Distributional shift in reinforcement learning (RL) is defined
inconsistently across the literature, through mismatches in trajectories,
visitation distributions, or training and deployment data. It is also
studied alongside out-of-distribution (OOD) generalization and
non-stationarity, yet how these three terms are related remains unclear.
Because these labels do not specify which part of the interaction changed,
methods are often described as robust to distributional shift, OOD
conditions, or non-stationarity after evaluation under only a few particular
changes, leaving the scope of such claims unclear. We develop a unified taxonomy that relates distributional shift, OOD
generalization, and non-stationarity by separating what changes (the
component-level source), when it changes (the interaction regime), and what
is observed (the resulting changes in interaction data and performance). We
complement it with an evaluation framework that measures how methods
respond to a specified shift, through immediate generalization and subsequent
adaptation. We factorize the joint distribution of a one-step interaction into five
components, each produced by its own generating function: initial-state
distribution, observation mechanism, policy implementation (distinct from the
policy being optimized), transition dynamics, and reward mechanism. A
distributional shift is a mismatch between the pre- and post-shift joint
distributions, which requires at least one generating function to change.
Three interaction regimes relate the two conditions: OOD generalization,
non-stationary adaptation, and their combination. A regime together with the
changed components specifies the shift. We illustrate the taxonomy with
published benchmark results, controlled Gridworld experiments with DQN, A2C,
and PPO, and a mapping of selected studies. For adaptation, the evaluation
framework combines pre- and post-shift performance, endpoint performance,
recovery time, and recovery gain into a shift adaptation score (SAS), and
reports immediate shift impact separately. Within the taxonomy, distributional shift is the joint mismatch caused by a
component-level change, OOD generalization and non-stationarity are the
regimes in which it occurs, and the discrepancies compared by existing
definitions are observed effects rather than shift sources. The selected
studies each evaluate only some components, and none in this selection tests
an observation-mechanism or policy-implementation shift. Across the
benchmarks and our experiments, method responses and rankings vary with the
shifted component, and no single method performs best under all of them. In
the Gridworld experiments, the SAS and its component metrics quantify these
differences in degradation and recovery. Distributional shift, OOD generalization, and non-stationarity are therefore
complementary descriptions of shifted interaction rather than separate
problems. Robustness claims should specify the interaction regimes,
components, and changes evaluated, since strong performance under one shift
does not establish robustness under others. Identifying an unknown
component-level source from observed interaction data remains a direction
for future research.

\end{abstract}

\section{Introduction}
\label{sec:introduction}

In reinforcement learning (RL), agent behaviour and experience are generated
through repeated interaction with the environment
\cite{suttonReinforcementLearningIntroduction2018}. At each interaction step,
the environment is in a state, the agent receives an observation and selects
an action, after which the environment transitions to a new state and
generates a reward. Consequently, trajectories, state--action visitation
distributions, return distributions, and agent performance are jointly
determined by the local generative components governing this interaction.

When one or more of these components change across interaction conditions, the
interaction data and agent performance may also change. The RL literature
studies such changes through three related but commonly separated framings.
First, \emph{distributional shift} is characterized through discrepancies in
interaction data, including trajectory distributions
\cite{luoMitigatingDistributionShift2025}, state--action visitation
distributions \cite{wangImprovingGeneralizationOffline2025}, and
offline--online data distributions
\cite{leeAddressingDistributionShift2020}. Second,
\emph{out-of-distribution (OOD) generalization} concerns differences between
training and evaluation or deployment conditions
\cite{korkmazSurveyGeneralizationDeep,
PackerAsessingGeneralization,
cobbeQuantifyingGeneralizationReinforcement2019}. Third,
\emph{non-stationarity} concerns temporal changes that occur during ongoing
interaction
\cite{padakandlaSurveyReinforcementLearning2022,
papoudakisDealingNonStationarityMultiAgent2019}.

These framings do not convey the same kind of information. Distributional
shift formulations identify a mismatch between two distributions over
interaction data, typically the data from which the agent learned and the
data it later encounters. Even these formulations compare different
distributions, such as trajectories, state--action visitation, or offline and
online data, so the term itself is not used consistently. OOD generalization
and non-stationarity, in contrast, characterize relations between the
interaction conditions that generate such data. Yet rather than being treated
as complementary views of what may be the same underlying change, the three
framings are frequently presented as separate RL problems, without an
explicit account of how they are related or which part of the
agent--environment interaction changed. This raises our first research question:
\begin{quote}
\textbf{RQ1:} \emph{How can distributional shift, OOD generalization, and
non-stationarity be related within a common structure of agent--environment
interaction?}
\end{quote}

A second issue concerns the scope of robustness claims. A method may be
described as robust to ``distributional shift'', ``OOD conditions'', or
``non-stationarity'' after being evaluated under only one or a few particular
disturbances. However, such evidence directly establishes robustness only
under the shift sources and conditions that were evaluated. For example,
robustness to visual variation does not by itself establish robustness to
changes in the initial-state distribution, transition dynamics, reward mechanism, or
policy implementation. Existing empirical evidence supports source-specific
evaluation. Robust-Gymnasium \cite{guROBUSTGYMNASIUMUNIFIED2025} evaluates
disruptions affecting different stages of agent--environment interaction and
demonstrates substantial weaknesses across the evaluated RL methods, while
RL-ViGen \cite{yuanRLViGenReinforcementLearning} evaluates multiple visual
generalization categories and reports that no single evaluated method performs
best across all tasks and categories. This raises our second research question:
\begin{quote}
\textbf{RQ2:} \emph{Does robustness demonstrated under one component-level
shift source justify a general robustness claim covering other, untested
sources within the same broad interaction regime?}
\end{quote}

To address RQ1, we propose a unified taxonomy for shifted
interaction in RL, starting from the joint distribution $p(s,o,a,s',r)$ of a
one-step agent--environment interaction. Its factorization provides a common structure that identifies five potential shift sources: the initial-state
distribution, observation mechanism, policy implementation, transition
dynamics, and reward mechanism. Within the policy component, we distinguish
the policy being optimized through learning from its implementation as a
potential source of shift.

The taxonomy separates three pieces of information: (1) which local generative
component changes, (2) under which interaction regime the change occurs, and
(3) what effects are observed in interaction data and agent performance. The
first two specify the shift source and its interaction regime, whereas the third
describes its possible consequences. Under this taxonomy, OOD generalization
and non-stationarity are interaction regimes, distinguished by whether the
change occurs between training and deployment, during ongoing interaction, or
both. Distributional shift corresponds to the resulting change in the one-step
joint distribution, while the discrepancies in trajectories, visitation
distributions, or returns emphasized by existing formulations are its observed
effects.

To address RQ2, we combine the interaction regime and the changed components
into a structurally specified shift, so that each robustness result is tied to
a specified component-level source and interaction regime. We then examine
published OOD benchmark results and controlled non-stationary Gridworld
experiments with DQN, A2C, and PPO, in which method responses can be compared
under known interventions in different components. To complement standard
performance measures such as return or success rate, we develop an evaluation
framework that measures the immediate impact of a shift and combines
performance before and after the shift, endpoint performance, recovery time,
and recovery gain into a shift adaptation score (SAS). A protocol guides how
the taxonomy and evaluation framework are applied.

Our objective is to specify shifted interaction and qualify robustness claims,
not to propose another robustness algorithm, disturbance benchmark, or method
for diagnosing an unknown shift source. In our controlled experiments, the
source is the component directly changed by the experimenter. In uncontrolled
settings, changes in interaction data or performance may indicate that the
interaction process has changed, but identifying the precise component-level
source remains a separate inverse problem and an open research direction.

\noindent\textbf{\textit{Our Contributions.}}
The contributions address the problems raised above as follows:
\begin{itemize}
\item To resolve the inconsistent use of distributional shift and clarify
its relation to OOD generalization and non-stationarity (RQ1), we formulate a unified taxonomy from the joint distribution of a
one-step agent--environment interaction. It identifies five potential
component-level shift sources, each produced by its own generating function,
and distinguishes policy implementation from the policy being optimized.
Within this taxonomy, distributional shift is the joint mismatch caused by a
component-level change, OOD generalization, non-stationary adaptation, and
their combination are the interaction regimes in which it occurs, and the
discrepancies compared by existing definitions are observed effects rather
than shift sources.

\item To make the scope of robustness claims explicit (RQ2),
we combine the interaction regime and the changed components into a
structurally specified shift. Evidence obtained under one such shift does
not by itself establish robustness under untested component--regime
combinations.

\item To support fair comparisons of methods under specified shifts, we
develop an evaluation framework that measures immediate shift impact and
combines performance before and after the shift, endpoint performance,
recovery time, and recovery gain into the SAS. We also provide a protocol
for applying the taxonomy and evaluation framework.

\item To provide evidence for both research questions, we illustrate the taxonomy and
evaluation framework through published OOD benchmark results, controlled
non-stationary Gridworld experiments with DQN, A2C, and PPO, and a mapping of
selected studies. They show that method responses and rankings vary across
the tested components, and that the selected studies evaluate only some
components.

\end{itemize}
\section{Background}

%====================================================
\subsection{Markov Decision Processes in Reinforcement Learning}
%====================================================

Reinforcement learning is commonly formulated through sequential
interaction between an agent and an environment. At each time step
$t$, the agent selects an action $A_t$, after which the environment
produces a reward $R_{t+1}$ and a subsequent state $S_{t+1}$.
In the finite Markov Decision Process (MDP) formulation, states,
actions, and rewards belong to the sets $\mathcal{S}$, $\mathcal{A}$,
and $\mathcal{R}$, respectively
\cite{Howard_DynamicProg_1960,2005PuttermanMDP,
suttonReinforcementLearningIntroduction2018}.

Following~\cite{suttonReinforcementLearningIntroduction2018},
the one-step dynamics are specified by
\begin{equation}
p(s',r\mid s,a)
=
\Pr\left\{
S_{t+1}=s',R_{t+1}=r
\mid S_t=s,A_t=a
\right\}
\label{eq:mdp_joint_dynamics}
\end{equation}
For every $s\in\mathcal{S}$ and $a\in\mathcal{A}$, this distribution
satisfies
\begin{equation}
\sum_{s'\in\mathcal{S}}
\sum_{r\in\mathcal{R}}
p(s',r\mid s,a)
=1
\label{eq:mdp_joint_normalization}
\end{equation}

In a fully observable MDP, a stationary policy specifies the
probability of selecting an action given the current state:
\begin{equation}
\pi(a\mid s)
=
\Pr\left\{
A_t=a\mid S_t=s
\right\}
\label{eq:state_based_policy}
\end{equation}
The learning objective is to find a policy that maximizes the
expected discounted return
\begin{equation}
G_t
=
\sum_{k=0}^{\infty}\gamma^k R_{t+k+1}
\label{eq:discounted_return}
\end{equation}
where $\gamma\in[0,1)$ is the discount factor. The value $\gamma=1$
may also be used in episodic tasks when the return is well-defined
\cite{suttonReinforcementLearningIntroduction2018}.

Under partial observability, the agent receives an observation
rather than directly accessing the complete environment state
\cite{kaelblingPlanningActingPartially1998}.
For the state-conditioned observation model considered here,
the observation mechanism is:
\begin{equation}
p(o\mid s)
=
\Pr\left\{
O_t=o\mid S_t=s
\right\}
\label{eq:observation_mechanism}
\end{equation}
The agent may use this observation together with information retained
from previous interactions to select an action.

% Whether actions are selected from states or observations, our work considers
% the policy from two perspectives: as the object optimized through
% learning to maximize expected return, and as an implemented mechanism
% that produces actions during interaction. The joint distribution of the
% interaction variables, introduced in
% Section~\ref{subsec:one_step_factorization}, provides the starting point
% for our unified taxonomy.

Our work considers the policy from two perspectives: as the object optimized
through learning to maximize expected return, and as the implemented mechanism
that produces actions during interaction. Both perspectives apply whether
actions are selected from states or observations.
Section~\ref{subsec:one_step_factorization} introduces the joint distribution
of the interaction variables, which provides the starting point for our
unified taxonomy.

%====================================================
\subsection{In-Distribution (ID) and Out-of-Distribution (OOD)}
\label{subsec:bg_ood}

Classical statistical learning assumes that training and evaluation samples are
generated under the same distribution. When
$P_{\text{train}}(X,Y)\neq P_{\text{deploy}}(X,Y)$, the
training--deployment relation violates this fixed-distribution assumption
\cite{quinonerocandelaDatasetShiftMachine2010}.
The term \emph{in-distribution} (ID) is commonly used when evaluation data are
consistent with the training distribution, whereas
\emph{out-of-distribution} (OOD) refers to evaluation under a distribution that
differs from the one represented during training.

In RL, the same idea is commonly used to study generalization across
environments or task conditions
\cite{korkmazSurveyGeneralizationDeep,PackerAsessingGeneralization,
cobbeQuantifyingGeneralizationReinforcement2019}.
In this paper, we treat OOD generalization as an \emph{interaction regime}:
it describes a mismatch between the generative conditions represented during training and those encountered during deployment or evaluation.
This label does not by itself determine what local mechanism changed.
For example, an OOD evaluation may be produced by a change in the initial-state
distribution, observation mechanism, transition dynamics, reward mechanism, or
a combination of these factors. The unified taxonomy introduced later is
used to specify that additional structural information.

Within our unified taxonomy, OOD retains its established meaning as a
difference between training and evaluation or deployment conditions. It
corresponds to the train--deployment relation considered in RQ1 (Section~\ref{sec:introduction}). The
component-level description complements this relation by specifying which
local generative component changed.

\subsection{Stationarity and Non-Stationarity}
\label{subsec:bg_nonstationarity}

In probability theory, a stochastic process
$\{Z_t\}_{t\geq 0}$ is strictly stationary when its finite-dimensional
distributions are invariant under time shifts
\cite{DoobStochastic1990}.
For any integer $k\geq 1$, time indices $t_1,\ldots,t_k$, and temporal
offset $\tau$,
\[
P(Z_{t_1},\ldots,Z_{t_k})
=
P(Z_{t_1+\tau},\ldots,Z_{t_k+\tau})
\]
If this invariance does not hold, the process is non-stationary.

For controlled Markov processes, a related distinction concerns whether the
transition and reward mechanisms are time homogeneous. A transition kernel
$p(s'\mid s,a)$ that remains unchanged over time is time homogeneous, whereas a
kernel $p_t(s'\mid s,a)$ that changes with $t$ is time dependent
\cite{1997_Norris_MarkovChain}.
RL research uses the broader notion of non-stationarity for interaction
settings in which relevant parts of the agent--environment process evolve
during ongoing interaction
\cite{padakandlaSurveyReinforcementLearning2022,
papoudakisDealingNonStationarityMultiAgent2019}.

Within the same unified taxonomy, non-stationarity retains its established
meaning as temporal change during ongoing interaction. It corresponds to the
temporal relation considered in RQ1 (Section~\ref{sec:introduction}).

% ==========================================================
\section{Related Work and Gap}
\label{sec:related_work}
% ==========================================================

Prior work describes distributional shift in reinforcement learning
at different levels. Some studies examine changes in particular
components, while others describe differences in trajectories,
state--action visitation, datasets, or performance. OOD generalization
and non-stationarity specify the interaction regime in which these
changes occur. These perspectives are complementary: a component-level
change can occur between training and evaluation or during ongoing
interaction, and can alter several performance and data distributions.

We review these perspectives to clarify how our taxonomy connects
shift components, interaction regimes, and observed effects, and
how the evaluation framework supports comparisons within this structure.

% ----------------------------------------------------------
\subsection{Shift-Related Surveys, Frameworks, and Benchmarks}
% ----------------------------------------------------------

Existing surveys organize RL research around zero-shot
generalization~\cite{kirkSurveyZeroshotGeneralisation2023},
state representation~\cite{echchahedSurveyStateRepresentation2025},
sim-to-real transfer~\cite{zhaoSimtoRealTransferDeep2020},
generalization in deep RL~\cite{korkmazSurveyGeneralizationDeep},
non-stationary environments~\cite{padakandlaSurveyReinforcementLearning2022},
and continual learning~\cite{khetarpalContinualReinforcementLearning2022}.
In multi-agent RL, changing policies of other agents are also studied
as a source of non-stationarity
\cite{papoudakisDealingNonStationarityMultiAgent2019}.
Their organizing criteria include evaluation settings, representation
objectives, sources of non-stationarity, and transfer methods.
Our taxonomy connects these perspectives by separating the component
from which a shift originates, the interaction regime, and the
resulting changes in data and performance.

A closely related structural framework is presented
in~\cite{haiderDomainShiftsReinforcement2021}.
It describes training--testing differences through the MDP elements
$\mathcal{S}$, $\mathcal{A}$, $P$, $R$, and $\mu_0$, relating them
to environmental safety, robustness, novelty, OOD conditions, and
domain adaptation. Our formulation instead starts from the joint
distribution $p(s,o,a,s',r)$ of a one-step interaction. Its factorization
explicitly represents the observation mechanism and allows policy
implementation to be distinguished from policy optimization.
We also relate the component-level sources to OOD generalization,
non-stationary adaptation, and combined settings.

Empirical studies examine generalization across training and testing
environments~\cite{zhangStudyOverfittingDeep2018,PackerAsessingGeneralization}.
Robust-Gymnasium introduces a Disrupted-MDP framework with disturbances
to observations, rewards, executed actions, and the environment
\cite{guROBUSTGYMNASIUMUNIFIED2025}.
RL-ViGen evaluates visual generalization across appearance, lighting,
camera views, scene structures, and embodiments, reporting that no
evaluated method performs best across all tasks and categories
\cite{yuanRLViGenReinforcementLearning}.
These benchmarks motivate evaluation across different changes and
show why robustness claims should specify the tested conditions.

Our contribution builds on these structural descriptions and benchmarks.
We organize distributional shifts by their component-level source and
interaction regime, while distinguishing these from downstream effects.
We additionally provide an application protocol and metrics for
evaluating immediate shift impact and adaptation through pre- and
post-shift performance and recovery.

% ----------------------------------------------------------
\subsection{Existing Definitions of Distributional Shift in RL}
% ----------------------------------------------------------

Several RL formulations describe distributional shift through
differences in trajectories, visitation distributions, or training
and execution data. These formulations explain important learning
problems, but the distributions being compared are not always
individual shift-source components.

\paragraph{Trajectory-ratio formulation.}
In model-based offline RL, distributional shift can be expressed
through a ratio between training and environment trajectory
distributions~\cite{luoMitigatingDistributionShift2025}.
For state--action trajectories with a shared initial-state distribution,
the ratio factorizes, wherever the denominator is nonzero, as:
\begin{equation}
\frac{q^{\pi_c}(\tau)}{p^\pi(\tau)}
=
\prod_t
\underbrace{
\frac{q(s_{t+1}\mid s_t,a_t)}
     {p(s_{t+1}\mid s_t,a_t)}
}_{\text{model bias}}
\underbrace{
\frac{\pi_c(a_t\mid s_t)}
     {\pi(a_t\mid s_t)}
}_{\text{policy shift}}
\label{eq:luo_ratio_cont}
\end{equation}
Here, $p^\pi(\tau)$ is generated by policy $\pi$ under environment
dynamics $p$, while $q^{\pi_c}(\tau)$ uses policy $\pi_c$ and
training dynamics $q$. If their initial-state distributions differ,
the ratio also includes that initial-state probability ratio.

This formulation separates discrepancies in dynamics models and
policies. However, model bias does not necessarily indicate that
the environment dynamics changed. Likewise, the policy discrepancy
does not itself establish a policy-implementation shift under our
taxonomy: differences between data-collection and learned policies
can arise through ordinary policy optimization.

\paragraph{State--action visitation mismatch.}
Offline RL also describes distributional shift as a mismatch between
the visitation distribution induced by a learned policy and the
distribution represented by its training dataset
\cite{wangImprovingGeneralizationOffline2025}.
A policy may select actions in regions poorly covered by the dataset,
even when the environment dynamics remain unchanged.
This describes a data-coverage problem rather than direct evidence
of a transition-dynamics shift.

\paragraph{Training--rollout distribution mismatch.}
In off-policy RL, the distribution used to train an action-value
function can differ from the visitation distribution subsequently
induced by the updated policy
\cite{chenForesightDistributionAdjustment2024}.
This relationship can be summarized as:
\begin{equation}
d^{\mathrm{train}}
\longrightarrow
Q_{k+1}
\xrightarrow{\ \mathrm{guide}\ }
\pi_{k+1}
\xrightarrow{\ \mathrm{rollout}\ }
\bigl(d^{\pi_{k+1}},\,\eta(\pi_{k+1})\bigr)
\label{eq:chen_flow_cont}
\end{equation}
The training distribution $d^{\mathrm{train}}$ is used to learn
$Q_{k+1}$, which guides the policy update. The updated policy then
induces visitation distribution $d^{\pi_{k+1}}$ and performance
$\eta(\pi_{k+1})$.
This mismatch arises through the learning process and does not alone
identify a change in the environment or policy implementation.

\paragraph{Offline--online dataset mismatch.}
An agent trained from an offline dataset may encounter a different
data distribution when it begins online interaction
\cite{leeAddressingDistributionShift2020}.
Differences in state--action coverage can produce unfamiliar samples
and unstable fine-tuning. Methods addressing this problem balance
or otherwise reconcile offline and online experience.
The mismatch concerns the collected data. Therefore, identifying a component-level
shift additionally requires establishing which generating mechanism
changed.

These formulations compare different probability objects and explain
how their discrepancies affect learning. Our taxonomy complements
them by specifying the component-level source and interaction regime
when a generating mechanism changes. Accordingly, data mismatch,
model error, and ordinary policy updates are not automatically
classified as environment or policy-implementation shifts.

% ----------------------------------------------------------
\subsection{Robustness and Adaptation Methods}
% ----------------------------------------------------------

Methods for non-stationary single-agent environments include
change detection and context-specific policy learning
\cite{padakandlaReinforcementLearningAlgorithm2020},
behaviour-based detection and adaptation
\cite{liuBehaviorAwareApproachDeep2024}, and context detection
\cite{dasilvaImprovingReinforcementLearning2006}.
Hybrid safe RL also examines online model updates following
changes encountered after offline training
\cite{hickmanHybridSafeReinforcement2025}.

Multi-agent methods address changes arising from other agents through
transition modelling~\cite{jiangI2QFullyDecentralized2022},
trust-region control~\cite{liDEALINGNONSTATIONARITYMARL2022},
smooth Q-learning~\cite{amhraouiSmoothQLearningAlgorithm2023},
multi-timescale updates~\cite{emamiNonStationaryPolicyLearning2023},
belief modelling~\cite{zhaiDynamicBeliefDecentralized2023},
partner modelling~\cite{xuDecentralizedMultiagentCooperation2024},
meta-learning~\cite{al-shedivatContinuousAdaptationMetaLearning2018},
and clustering~\cite{duMultiagentReinforcementLearning2024}.
The component-level interpretation of these changes depends on
whether the interaction is represented from an individual agent's
perspective or jointly across agents.

Methods for generalization and adaptation include
meta-reinforcement learning
\cite{beckSurveyMetaReinforcementLearning2024}, with approaches
such as MAML~\cite{finnModelAgnosticMetaLearningFast2017},
RL$^2$~\cite{duanRL$^2$FastReinforcement2016},
DiAMetR~\cite{ajayDistributionallyAdaptiveMeta2022}, and
PSBL~\cite{xuMetaReinforcementLearningRobust2024}.
Under our operational distinction, adaptation includes using post-shift
experience to learn task-specific behaviour through interaction history, as
in RL$^2$ and PSBL, even when network parameters remain fixed. Adaptation of
this kind can occur within an OOD regime. Because the change in these
protocols occurs between training and evaluation,
we classify them as OOD.

Domain randomization broadens training conditions over visual
or physical parameters
\cite{tobinDomainRandomizationTransferring2017}.
Examples include visual domain randomization
\cite{slaouiRobustVisualDomain2020},
dynamics randomization
\cite{muratoreNeuralPosteriorDomain2021,
tiboniDROPOSimtorealTransfer2023},
and randomization followed by real-world fine-tuning
\cite{shakerimovEfficientSimtoRealTransfer2023}.
Other approaches include CROP
\cite{altmannCROPDistributionalShiftRobust2023},
robust transition modelling
\cite{herremansRobustModelBasedReinforcement2024},
evaluation-time policy switching
\cite{neggatuEvaluationTimePolicySwitching2025},
distribution-aware value learning
\cite{chenForesightDistributionAdjustment2024},
and latent distribution alignment
\cite{wangImprovingGeneralizationOffline2025}.

These methods address different changes and learning settings.
The mechanism used to improve robustness does not necessarily
identify the source of the evaluated shift: for example, modifying
observation processing can address a change in scene configuration.
Table~\ref{tab:algo-taxonomy} therefore maps selected experimental
protocols by the change introduced and whether learning continues
under the changed conditions.

Our taxonomy makes the scope of these evaluations explicit.
Robustness demonstrated under one component-level change leaves
performance under other changes to be examined.
The accompanying protocol and metrics support evaluation of
generalization and adaptation within the specified scope.

% ----------------------------------------------------------
\subsection{Gap and Our Position}
\label{subsec:gap_position}
% ----------------------------------------------------------

\begin{figure*}[t]
    \centering
    \includegraphics[width=\textwidth]{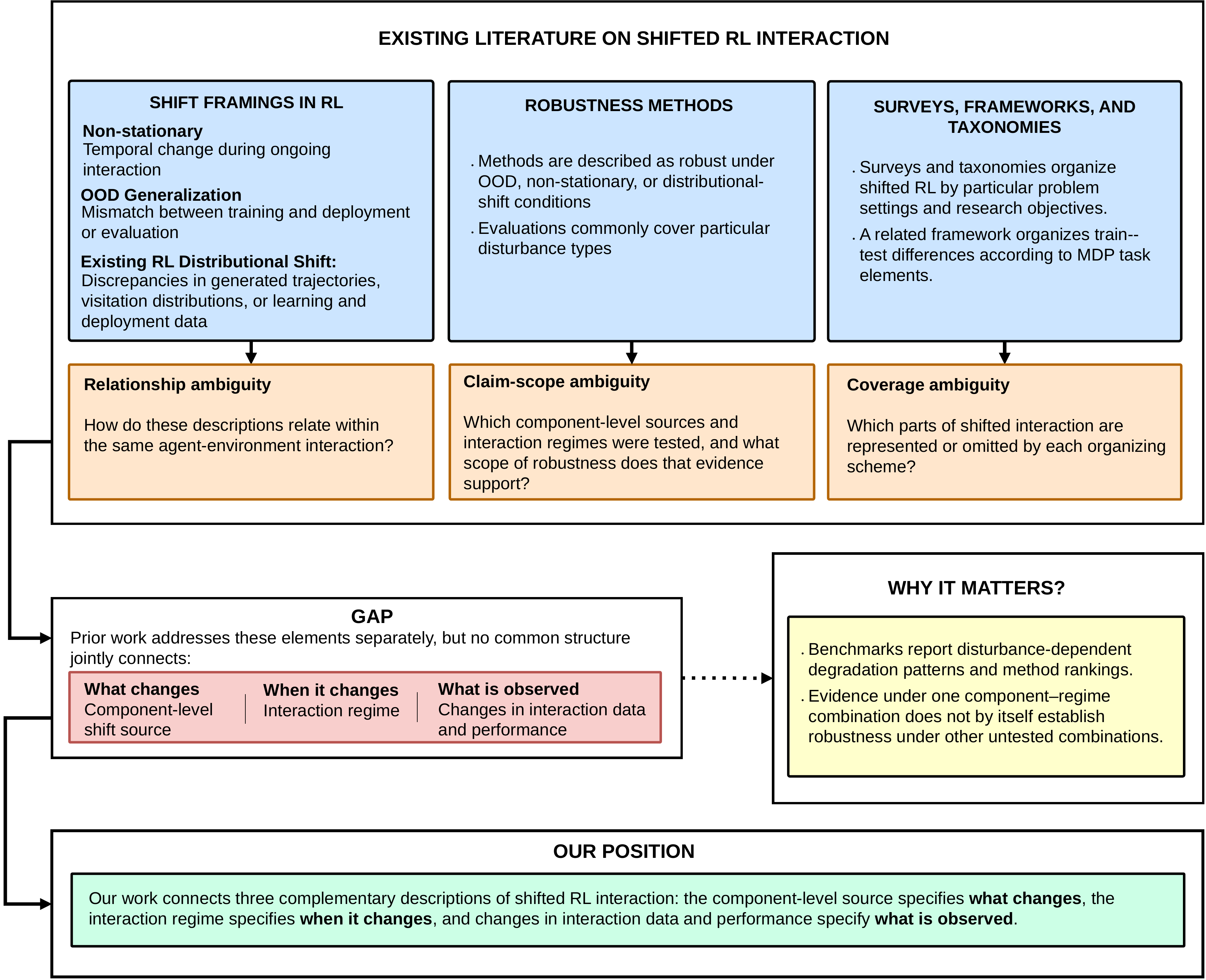}
    \caption{The gap identified from existing literature and the position of
    this work. Existing shift framings, robustness methods, and
    surveys, frameworks, and taxonomies lead respectively to relationship,
    claim-scope, and coverage ambiguities. The identified gap concerns the
    absence of a common structure connecting what changes, when it changes,
    and what is observed. The dotted arrow indicates why this gap matters, while the solid arrows show the main argumentative flow toward our position. Neither represents temporal order.}
    \label{fig:gap_position}
\end{figure*}

The literature reviewed above approaches shifted RL interaction from three
main directions, as summarized in Figure~\ref{fig:gap_position}. First,
existing shift framings characterize non-stationarity as temporal change, OOD
generalization as training--deployment mismatch, and distributional shift
through discrepancies in generated interaction data. Their relationship within
the same agent--environment interaction, however, remains unclear. Second,
robustness methods are often described using these broad shift labels, although
their evaluations cover particular disturbance conditions. This creates
ambiguity regarding which component-level sources and interaction regimes are
actually supported by the reported evidence. Third, existing surveys,
frameworks, and taxonomies organize shifted RL according to different problem
settings, research objectives, or task elements, leaving differences in which
parts of shifted interaction each scheme represents.

The individual concepts represented in Figure~\ref{fig:gap_position} are not
claimed to be new. Rather, the gap is that the reviewed literature addresses
them separately without a common structure that jointly connects the
component-level shift source (\emph{what changes}), the interaction regime
(\emph{when it changes}), and the resulting changes in interaction data and
performance (\emph{what is observed}). Without this connection, the empirical
scope of a broad robustness claim cannot be determined clearly from the shift
label alone.

This gap matters because existing benchmark evidence indicates that robustness
depends on the disturbance being evaluated. Robust-Gymnasium reports different
degradation patterns across disturbance types
\cite{guROBUSTGYMNASIUMUNIFIED2025}, while RL-ViGen shows that method
performance varies across visual-generalization categories, with no single
evaluated method consistently performing best
\cite{yuanRLViGenReinforcementLearning}. Evidence obtained under one
component--regime combination therefore does not by itself establish
robustness under other untested combinations.

Our position is that the component-level shift source, interaction regime, and
resulting changes provide three complementary descriptions of shifted RL
interaction. The component-level source specifies what changes, the interaction
regime specifies when it changes, and changes in interaction data and
performance specify what is observed. We connect these descriptions within a
unified taxonomy while preserving the established meanings of
distributional shift, OOD generalization, and non-stationarity. The next section
formalizes this taxonomy and develops its implications in detail.

% ==========================================================
\section{A Unified Taxonomy of Distributional Shifts in Reinforcement Learning}
\label{sec:unified_taxonomy}
% ==========================================================

Following the position established in
Section~\ref{subsec:gap_position}, we characterize distributional shift in RL
through three complementary descriptions: 
the component-level taxonomy specifies what changes, the interaction regime specifies when it changes, and the resulting changes describe what is subsequently observed in interaction data or agent
performance.

\begin{figure}[h]
\centering
\includegraphics[width=0.95\textwidth]{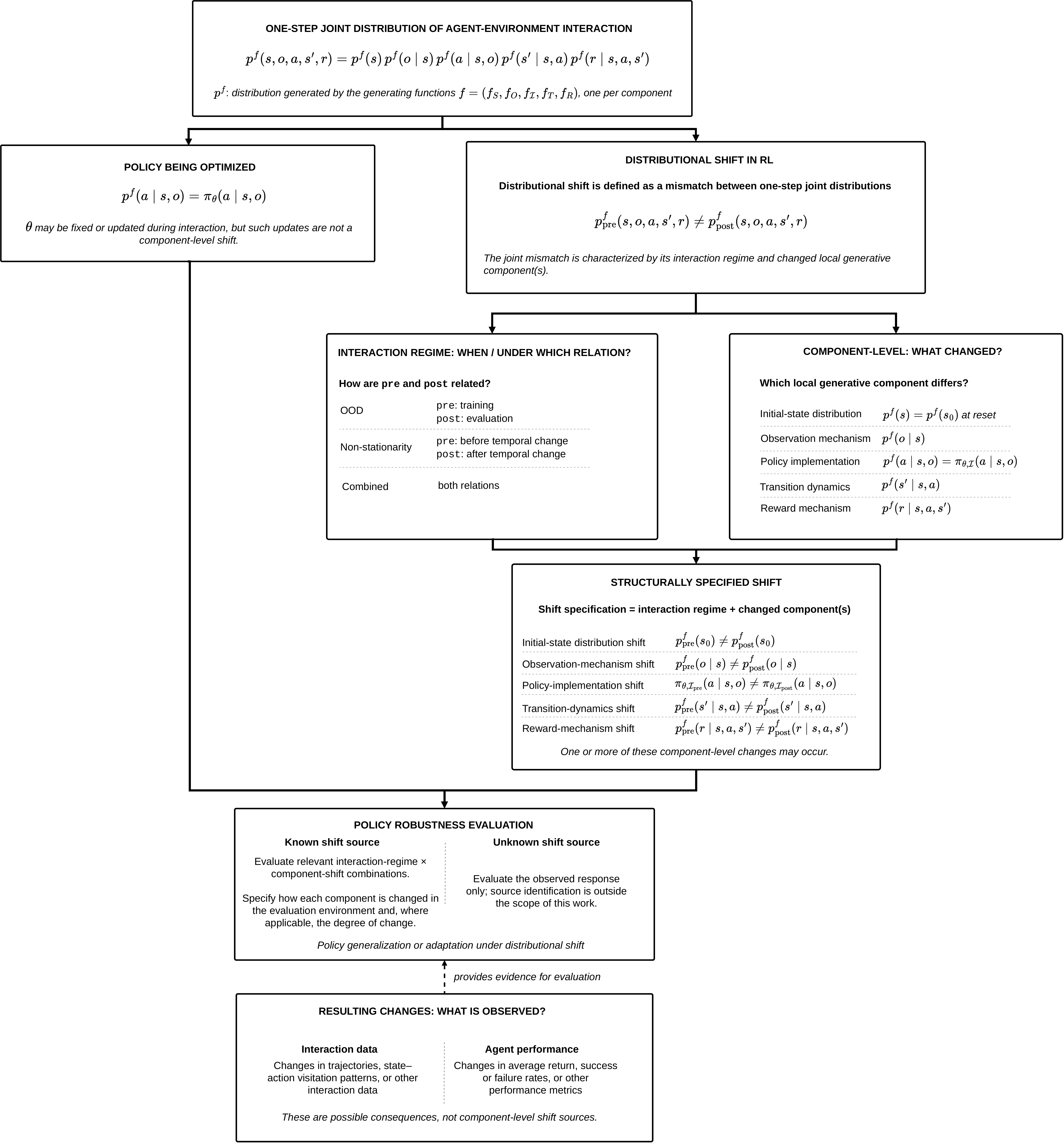}
\caption{Unified taxonomy of distributional shifts in reinforcement
learning and its implications for robustness evaluation.}
\label{fig:taxonomy_flow}
\end{figure}

Figure~\ref{fig:taxonomy_flow} summarizes the taxonomy. It is read from top
to bottom. The top box represents the one-step joint distribution $p^f$ generated by
the generating functions $f$ of the agent--environment interaction. This distribution contains both the
policy whose robustness is being evaluated and the local mechanisms through
which a shift may arise. The first branch therefore identifies the policy
under evaluation, whereas the second branch defines distributional shift by
comparing the joint interaction distributions under pre- and post-shift
conditions.

Within the distributional-shift branch, the interaction regime and the
component-level taxonomy provide two complementary descriptions. The
interaction regime specifies how the pre- and post-shift conditions are
related, while the component-level taxonomy identifies which local mechanism
differs between them. Their combination produces a structurally specified
shift. This shift specification is then paired with the policy under
evaluation to assess generalization or adaptation. Finally, changes in
interaction data and performance are treated as possible consequences of the
shift, rather than as additional component-level sources.

% ----------------------------------------------------------
\subsection{One-Step Joint Interaction Factorization}
\label{subsec:one_step_factorization}
% ----------------------------------------------------------

% We begin with the joint distribution of the variables generated within one
% agent--environment interaction step. These variables are produced by the
% generating functions $f=(f_S,f_O,f_{\mathcal{I}},f_T,f_R)$, one for each
% component: $f_S$ generates the initial state $s$ and scene configuration at
% reset, $f_O$ the observation $o$ of a state, $f_{\mathcal{I}}$ the executed
% action $a$ through the policy implementation, $f_T$ the next state $s'$, and
% $f_R$ the reward $r$. We write $p^f$ for the distribution they generate.  A component-level
% shift refers to a change in one of these functions that alters the distribution it generates, and differences in the
% generated data are its observable consequence. For the variable ordering $(s,o,a,s',r)$,
% the chain rule of probability gives

We begin with the joint distribution of the variables generated within one
agent--environment interaction step. These variables are produced by the
generating functions $f=(f_S,f_O,f_{\mathcal{I}},f_T,f_R)$, one for each
component. Each function determines one distribution: $f_S$ the distribution of the initial state $s_0$ at reset, $f_O$ the distribution of the
observation $o$ given the state, $f_{\mathcal{I}}$ the distribution of the
executed action $a$ given the state and observation, $f_T$ the distribution
of the next state $s'$ given the current state and action, which describes
how the state evolves from one step to the next, and $f_R$ the distribution
of the reward $r$ given the transition. We write $p^f$ for the joint
distribution they generate. The distributions of the data observed over an
interaction depend on several of these functions together. For the variable
ordering $(s,o,a,s',r)$, the chain rule of probability gives

\begin{equation}
\begin{aligned}
p^f(s,o,a,s',r)
={}&
p^f(s)\,
p^f(o\mid s)\,
p^f(a\mid s,o)\,
p^f(s'\mid s,o,a)\,
p^f(r\mid s,o,a,s')
\end{aligned}
\label{eq:general_joint_factorization}
\end{equation}

This factorization follows the one-step interaction order: the environment is
in state \(s\), an observation \(o\) is generated from that state, an action
\(a\) is selected, the environment transitions to state \(s'\), and reward
\(r\) is generated from the resulting transition.

The chain-rule factorization in
Eq.~\ref{eq:general_joint_factorization} can be simplified according to the
conditional-dependence structure of the RL interaction.

First, under the Markov property, the next state is determined by
the current state and selected action. Once \(s\) and \(a\) are given, the
current observation \(o\) provides no additional information for determining
\(s'\). Therefore,

\begin{equation}
p^f(s'\mid s,o,a)
=
p^f(s'\mid s,a)
\label{eq:transition_reduction}
\end{equation}

Second, the reward mechanism is defined from the current state, selected
action, and resulting next state. Once \(s\), \(a\), and \(s'\) are given,
the current observation does not independently affect the generated reward.
Therefore,

\begin{equation}
p^f(r\mid s,o,a,s')
=
p^f(r\mid s,a,s')
\label{eq:reward_reduction}
\end{equation}

Applying these two conditional-independence relations gives the one-step
factorization used throughout the taxonomy:

\begin{equation}
\begin{aligned}
p^f(s,o,a,s',r)
={}&
p^f(s)\,
p^f(o\mid s)\,
p^f(a\mid s,o)\,
p^f(s'\mid s,a)\,
p^f(r\mid s,a,s')
\end{aligned}
\label{eq:one_step_joint_factorization}
\end{equation}

At the first step of an episode, \(s=s_0\), and the state factor is the
initial-state distribution \(p^f(s_0)\) generated by \(f_S\). At every later
step, the current state \(s\) is the next state \(s'\) of the previous step,
so it is produced by \(f_T\) and the preceding interaction rather than by
\(f_S\). Over an episode, the state therefore enters as a separate factor only
at reset, whereas the other four factors are produced by \(f_O\),
\(f_{\mathcal{I}}\), \(f_T\), and \(f_R\) at every step.

The action-selection term \(p^f(a\mid s,o)\) remains generic at this stage. Its
relation to the policy being evaluated and the policy implementation is
introduced separately in the subsequent component-level formulation. A separate next-observation variable \(o'\) is not required in this one-step
representation. At the following interaction step, \(o'\) is generated from
\(s'\) through the same observation mechanism, \(p^f(o'\mid s')\). Including
\(o'\) would therefore extend the interaction into the next step rather than
introduce a distinct one-step component.

Based on this factorization, the taxonomy proceeds through two connected branches, as shown in Figure~\ref{fig:taxonomy_flow}. The first identifies the policy being optimized through \(p^f(a\mid s,o)\), while the second compares the full one-step joint distributions under the pre- and post-shift conditions. This separates the policy being evaluated from the distributional shift under which it is evaluated.

% ----------------------------------------------------------
\subsection{Policy Being Optimized}
\label{subsec:policy_being_optimized}
% ----------------------------------------------------------

The action-selection factor in
Eq.~\ref{eq:one_step_joint_factorization} is represented by

\begin{equation}
p^f(a\mid s,o)
=
\pi_{\theta}(a\mid s,o)
\label{eq:policy_being_optimized}
\end{equation}

where \(\theta\) denotes the parameters of the policy being optimized.
These parameters may be fixed or updated during interaction, but such updates
are not considered a component-level shift and therefore do not define pre-
and post-shift situations. Under partial observability, the conditioning on \(s\) does not imply that the state is directly available to the agent, and any observation history, belief
state, or recurrent state used for action selection remains internal to the
policy. This branch identifies the policy, or evolving sequence of policies,
whose robustness is later evaluated under the specified shift.

% ----------------------------------------------------------
\subsection{Distributional Shift at the One-Step Joint-Distribution Level}
\label{subsec:joint_distribution_shift}
% ----------------------------------------------------------

The second branch defines distributional shift as a mismatch between the
one-step joint distributions of the pre- and post-shift situations:

\begin{equation}
p^f_{\mathrm{pre}}(s,o,a,s',r)
\neq
p^f_{\mathrm{post}}(s,o,a,s',r)
\label{eq:joint_interaction_shift}
\end{equation}

This inequality indicates that the joint distribution of agent--environment
interaction differs between the two situations. Here,
\(p^f_{\mathrm{pre}}\) and \(p^f_{\mathrm{post}}\) are the distributions
generated under the two situations for given policy parameters \(\theta\)
introduced in Section~\ref{subsec:policy_being_optimized}. A mismatch
therefore requires the generating function to differ for at least one
component, whereas differences in observed data that arise while the
generating function is unchanged, for example through learning updates of
\(\theta\), are not distributional shifts. A shift is thus defined by the
generating function that produces the interaction, not by how the generated
data appear.

As an illustration, consider obstacles that move by one cell at every step
under a fixed rule. The resulting states and observations differ from step to
step, but no shift occurs because the generating function is unchanged.
Replacing this rule with random or two-cell moves changes the transition
dynamics and therefore constitutes a shift. Section~\ref{sec:empirical_evidence}
illustrates this distinction in detail based on empirical evidence.

At this stage, \(\mathrm{pre}\) and \(\mathrm{post}\) only identify the
distributions being compared. How these situations are related and which
component or components produce the mismatch are specified in the subsequent
subsections through the interaction regime and component-level taxonomy.

% ----------------------------------------------------------
\subsection{Interaction Regime: How Pre and Post Are Related}
\label{subsec:interaction_regime}
% ----------------------------------------------------------

The interaction regime assigns meaning to the
\(\mathrm{pre}\) and \(\mathrm{post}\) indices in
Eq.~\ref{eq:joint_interaction_shift}. It specifies when, or under which
relation, the two interaction conditions are compared.

\begin{enumerate}

\item \textit{OOD regime.}
The pre-shift condition represents the interaction structure encountered or
represented during training, whereas the post-shift condition represents the
structure encountered during evaluation or deployment:

\begin{equation}
p^f_{\mathrm{pre}}
=
p^f_{\mathrm{train}}
\qquad
p^f_{\mathrm{post}}
=
p^f_{\mathrm{eval}}
\label{eq:ood_regime_relation}
\end{equation}

\item \textit{Non-stationary regime.}
The pre- and post-shift distributions describe conditions before
and after a temporal change during ongoing interaction:
\begin{equation}
p^f_{\mathrm{pre}}=p^f_{t^-}
\qquad
p^f_{\mathrm{post}}=p^f_{t^+}
\label{eq:nonstationary_regime_relation}
\end{equation}
where $t^-$ and $t^+$ denote the times immediately before and after
the change. 

\item \textit{Combined regime.}
Both relations are present: the temporal change may
occur during training, during evaluation, or during both. The relevant
distributions may therefore be represented as

\begin{equation}
p^f_{\mathrm{train},t}
\qquad
p^f_{\mathrm{eval},t}
\label{eq:combined_regime_relation}
\end{equation}

with temporal comparisons such as

\begin{equation}
p^f_{\mathrm{train},t^-}
\neq
p^f_{\mathrm{train},t^+}
\qquad \text{and/or} \qquad
p^f_{\mathrm{eval},t^-}
\neq
p^f_{\mathrm{eval},t^+}
\label{eq:combined_temporal_relation}
\end{equation}

The train--evaluation relation and the within-phase temporal relations can
therefore be examined separately within the same combined regime.

\end{enumerate}

Accordingly, OOD and non-stationarity identify how the compared conditions are
related. They do not, by themselves, identify which part of the interaction
process changes. The same component-level change may occur across training
and evaluation, during ongoing interaction, or under both relations.

% ----------------------------------------------------------
\subsection{Component-Level: What Changes}
\label{subsec:component_level_taxonomy}
% ----------------------------------------------------------

After the interaction regime establishes how the two situations are related,
let \(j\in\{\mathrm{pre},\mathrm{post}\}\) identify either situation, and
let \(p^f_j\) denote a distribution generated by the generating function
of situation \(j\). Indexing the factors of
Eq.~\ref{eq:one_step_joint_factorization} by \(j\) gives

\begin{equation}
\begin{aligned}
p^f_j(s,o,a,s',r)
={}&
p^f_j(s)\,
p^f_j(o\mid s)\,
p^f_j(a\mid s,o)\,
p^f_j(s'\mid s,a)\,
p^f_j(r\mid s,a,s')
\end{aligned}
\label{eq:condition_indexed_factorization}
\end{equation}

The component-level taxonomy identifies five local generative components,
each produced by its own generating function. The state factor enters as the
initial-state distribution at reset, and the other four as the conditional
factors of Eq.~\ref{eq:condition_indexed_factorization}:

\begin{enumerate}
  \item the initial-state distribution \(p^f_j(s)=p^f_j(s_0)\) at reset,
    \item the observation mechanism \(p^f_j(o\mid s)\),
    \item the policy implementation
    \(
    p^f_j(a\mid s,o)
    =
    \pi_{\theta,\mathcal{I}_j}(a\mid s,o)
    \),
    \item the transition dynamics \(p^f_j(s'\mid s,a)\), and
    \item the reward mechanism \(p^f_j(r\mid s,a,s')\).
\end{enumerate}

For the policy-implementation factor, \(\theta\) denotes the policy parameters
introduced in Section~\ref{subsec:policy_being_optimized}, while
\(\mathcal{I}_j\) denotes the implementation condition under situation
\(j\). Thus, the policy parameters being optimized remain separate from
the condition-indexed implementation through which their action-selection
distribution is produced.

% ----------------------------------------------------------
\subsection{Structurally Specified Shift}
\label{subsec:structurally_specified_shift}
% ----------------------------------------------------------

A structurally specified shift combines the two dimensions introduced above.
The interaction regime specifies how the pre- and post-shift conditions are
related, whereas the component-level taxonomy specifies which generating function
differs between those conditions. Accordingly,

\begin{equation}
\text{shift specification}
=
\text{interaction regime}
+
\text{changed component(s)}.
\label{eq:structurally_specified_shift}
\end{equation}

Once the interaction regime assigns meaning to the
\(\mathrm{pre}\) and \(\mathrm{post}\) indices, the factors in
Eq.~\ref{eq:condition_indexed_factorization} produce the following five possible
component-level shifts. The comparisons are understood over the relevant
interaction support.

\begin{enumerate}

\item \textit{Initial-state distribution shift:}

\begin{equation}
p^f_{\mathrm{pre}}(s_0)
\neq
p^f_{\mathrm{post}}(s_0)
\label{eq:state_distribution_shift}
\end{equation}

This shift represents a difference in the generating function \(f_S\), which
generates the state at reset, such as the agent's starting position or the
scene configuration. In an OOD regime, the comparison may concern the
initial-state distributions used during training and evaluation. In a
non-stationary regime, it may concern the initial-state distributions before
and after a temporal change. Differences in the states visited after reset are
not initial-state distribution shifts. They are consequences of learning when
they arise only from changes in \(\theta\), including the coverage gap between
an offline dataset and a learned policy, and consequences of another component
when they arise, for example, from a change in \(f_T\).

\item \textit{Observation-mechanism shift:}

\begin{equation}
p^f_{\mathrm{pre}}(o\mid s)
\neq
p^f_{\mathrm{post}}(o\mid s)
\label{eq:observation_mechanism_shift}
\end{equation}

This shift represents a difference in how the same underlying state is mapped
to an observation under the compared conditions. It may involve changes in
sensing, representation, resolution, noise, corruption, or other properties
of the observation-generation mechanism.

\item \textit{Policy-implementation shift:}

\begin{equation}
\pi_{\theta,\mathcal{I}_{\mathrm{pre}}}(a\mid s,o)
\neq
\pi_{\theta,\mathcal{I}_{\mathrm{post}}}(a\mid s,o)
\label{eq:policy_implementation_shift}
\end{equation}

This shift occurs when a change in the implementation condition, such as
quantization, numerical precision, pruning, or hardware execution, alters the
action-selection distribution. The parameters \(\theta\) may remain fixed or
change through learning, but they are not indexed by the pre- and post-shift
conditions, so the shift source represented here is the change in
\(\mathcal{I}\).

\item \textit{Transition-dynamics shift:}

\begin{equation}
p^f_{\mathrm{pre}}(s'\mid s,a)
\neq
p^f_{\mathrm{post}}(s'\mid s,a)
\label{eq:transition_dynamics_shift}
\end{equation}

This shift represents a difference in how the environment transitions from
the same current state after the same selected action under the compared
conditions. The difference may occur between training and evaluation
environments, before and after a temporal change, or under both relations.

\item \textit{Reward-mechanism shift:}

\begin{equation}
p^f_{\mathrm{pre}}(r\mid s,a,s')
\neq
p^f_{\mathrm{post}}(r\mid s,a,s')
\label{eq:reward_mechanism_shift}
\end{equation}

This shift represents a difference in how rewards are generated for the same
current state, selected action, and resulting next state under the compared
conditions.

\end{enumerate}

% Within an interaction regime, one, two, or more component-level changes may
% occur. Because each factor in Eq.~\ref{eq:condition_indexed_factorization}
% is generated by its own function, a component-level change over the relevant
% interaction support changes the joint distribution in
% Eq.~\ref{eq:joint_interaction_shift}; conversely, a joint mismatch implies that at least one generating function
% has changed.

Within an interaction regime, one or more components may change. Because each component is generated by its own function, a change in any component alters the joint distribution in
Eq.~\ref{eq:joint_interaction_shift}, provided that the change affects
situations that actually occur during interaction. Conversely, if the joint
distributions differ, at least one generating function must have changed.

Changes in several generating functions cannot cancel each other out in the
joint distribution. Each function generates a different part of the
interaction. A change in one function therefore cannot be
undone by a change in another. For example, if \(f_S\) or \(f_T\) makes some states occur
more often, no change in \(f_O\) can restore how often those states occur,
because \(f_O\) only determines how each state is observed. As a result,
whenever a generating function changes in situations that occur, the joint
distribution changes, even if other functions change as well.

Changes can appear to cancel only when part of the interaction is hidden.
Consider two states, left and right. Before the shift, the agent starts on the
left with probability $0.7$ and its camera shows the true side. After the
shift, it starts on the left with probability $0.3$, and the camera is
mirrored, so the right side looks like the left. The same interaction then
gives different answers depending on what is observed:
\begin{itemize}
    \item \textit{Observations only.} Before the shift, ``left'' is observed
    with probability $0.7$, when the agent is on the left. After the shift,
    ``left'' is again observed with probability $0.7$, now when the agent is
    on the right. The distribution of observations is therefore unchanged.
    \item \textit{States and observations jointly.} Before the shift, the
    pair (left, ``left'') has probability $0.7$. After the shift, this pair
    has probability $0$, and the pair (right, ``left'') has probability $0.7$
    instead. The joint distribution has therefore changed.
    
   %  \item \textit{Observations only.} The observation ``left'' occurs with
   %  probability $0.7$ both before and after the shift, so the distribution of
   %  observations is unchanged.
   % \item \textit{States and observations jointly.} The pair (left, ``left'') has
   %  probability $0.7$ before the shift but $0$ after it, while the pair
   %  (right, ``left'') has probability $0.7$ instead, so the joint distribution
   %  has changed.
\end{itemize}
This is therefore a distributional shift under
Eq.~\ref{eq:joint_interaction_shift}, since both \(f_S\) and \(f_O\) have
changed, but it may not be detectable from the agent's observations alone
until the agent acts, after which its mirrored movements can reveal the
change. Identifying the changed functions in such cases from observed data is
left for future work (Section~\ref{sec:conclusion}).

% ----------------------------------------------------------
\subsection{Policy Robustness Evaluation}
\label{subsec:policy_robustness_evaluation}
% ----------------------------------------------------------

As illustrated in Figure~\ref{fig:taxonomy_flow}, policy robustness is
evaluated under a distributional shift. Depending on the objective, the
evaluation may concern a single policy that generalizes across interaction
conditions or a sequence of condition-specific policies developed during
adaptation.

The use of the structural taxonomy depends on whether the shift source is
known:

\begin{itemize}
    \item \textit{Controlled setting (known shift source).}
    When the interaction regime and changed components are known, the shift
    can be structurally specified. Evaluation should aim to cover the relevant
    combinations of interaction regimes and component-level changes. For each
    component tested, the changes evaluated and, where applicable, their degree
    should be specified for the evaluation environment. If only a subset of
    the relevant conditions is tested, the evaluated scope should be stated
    explicitly and robustness claims limited to that scope.

    \item \textit{Uncontrolled setting (unknown shift source).}
    When only resulting changes in interaction data or performance are
    observed, the component-level taxonomy cannot be used to determine the
    underlying shift source or support source-specific robustness claims.
    Identifying that source from its observed consequences is outside the scope
    of this work and remains an open research problem.
\end{itemize}

We also propose an evaluation framework that can be applied in
both controlled and uncontrolled settings because it evaluates the observed
response to the shift without requiring its component-level source to be
known. The framework is described in detail in
Section~\ref{sec:metric-shift}.

% ----------------------------------------------------------
\subsection{Resulting Changes: What Is Observed}
\label{subsec:resulting_changes}
% ----------------------------------------------------------

A distributional shift at the one-step joint-distribution level may propagate
to longer-horizon interaction quantities. As summarized at the bottom of
Figure~\ref{fig:taxonomy_flow}, the resulting changes fall into two broad
groups.

First, interaction data may change, including trajectory distributions,
state--action visitation distributions, and return distributions. Second,
agent performance may change, including average return, success or failure
rate, and other task-specific metrics. These quantities are possible
consequences of the shift, not additional component-level sources.

The observed effect of a distributional shift depends on its component-level
source. As demonstrated in the following section, different shift sources can
produce distinct patterns of interaction change, performance degradation, and
method ranking. These resulting changes provide the evidence used for
policy robustness evaluation and the proposed adaptation metric, while
remaining consequences rather than component-level sources.

%%%%%
%%%%%%

%%%%%%%%%%%%
%%%%%%%%%%%%
%%%%%%%%%%%

% % ==========================================================
\section{Empirical Evidence Across Distributional-Shift Settings}
\label{sec:empirical_evidence}
% ==========================================================

Our taxonomy in Figure~\ref{fig:taxonomy_flow} distinguishes
distributional shifts by their interaction regime and component-level
origin. We examine robustness through generalization under OOD
conditions and adaptation under non-stationarity. Evaluation should
cover the relevant combinations of regimes and shifted components,
including different changes and, where applicable, their degrees.
Robustness claims should remain limited to the conditions evaluated.

We examine these distinctions using existing studies and our own
controlled experiments. For OOD generalization, we draw on robustness
benchmarks
\cite{guROBUSTGYMNASIUMUNIFIED2025,yuanRLViGenReinforcementLearning}
that report performance under controlled training--evaluation
mismatches. We organize their results according to our taxonomy and
compare how methods perform across shifted components and the
changes tested within each component. For adaptation under non-stationarity, we conduct controlled Gridworld
experiments in which conditions change during ongoing learning.
We specify the timing and component-level source of each shift and
evaluate subsequent performance and recovery. 

For brevity, this section, including the tables, refers to components by
their unindexed factors, such as $p(s_0)$ or $p(o\mid s)$.

\subsection{Distributional Shifts in the OOD Generalization Setting: Evidence from Benchmark Literature}
\label{subsec:ood_evidence}

We examine selected results from RL-ViGen
\cite{yuanRLViGenReinforcementLearning} and Robust Gymnasium
\cite{guROBUSTGYMNASIUMUNIFIED2025} using the component-level definitions
in Section~\ref{subsec:component_level_taxonomy}. The comparison considers
performance variation both across shift components and across the changes
evaluated within a component.

RL-ViGen evaluates trained policies zero-shot under conditions that differ
from their training environments. The selected observation and action
experiments from Robust Gymnasium follow the same protocol of nominal training followed by shifted evaluation. %nominal-training-to-shifted-evaluation protocol 
Its reward experiment instead compares policies
trained with different reward disturbances and subsequently evaluated without
disturbances. We therefore discuss that experiment separately.

For RL-ViGen, performance degradation is calculated relative to each
method's positive training reference:
\begin{equation}
    \text{Performance degradation}
    =
    \frac{
        \text{Training reference}
        -
        \text{Evaluation score}
    }{
        \text{Training reference}
    }
    \times 100\%
    \label{eq:relative_performance_degradation}
\end{equation}
Lower percentages indicate smaller relative losses, and negative percentages
indicate improvement relative to the reference. Raw scores are retained
because a smaller relative loss does not necessarily imply higher
evaluation performance.

The RL-ViGen training references are obtained from the published
learning curves, which we digitized using WebPlotDigitizer
\cite{zotero-item-1524}, taking each reference as the mean performance
over the final $10\%$ of training, aggregated over the corresponding tasks. The resulting
percentages are approximate, reference-relative comparisons rather than
precise degradation estimates from matched clean and shifted evaluations.
For the observation and action experiments in Robust Gymnasium, the same
calculation uses the reported clean-evaluation reference in place of the
training reference. We interpret these comparisons descriptively and do
not infer statistical significance from the tabulated means.

The component mappings follow the state representation adopted in our
taxonomy. In particular, static appearance and layout fixed at reset are treated as part
of the initial state, appearance or lighting that changes during an episode as
part of the transition dynamics, and camera parameters as part of the
observation mechanism. Conditions combining several changes are interpreted
accordingly, but their aggregate scores do not isolate the contribution of
each component. Benchmark difficulty labels describe environment-specific
configurations and do not define a common degree of shift across components
or environments.

% ==========================================================
\subsubsection{Adroit}
% ==========================================================

Policies trained under the original Adroit scene and camera configuration
are evaluated under Color, Video, and Camera View changes.
Under our representation, static changes to background appearance, object
color, and table texture correspond to $p(s_0)$. Dynamic backgrounds and
moving illumination introduce transition-related changes,
$p(s'|s,a)$, while changes to camera position, orientation, and field of
view correspond to $p(o|s)$.

\begin{table*}[t]
\centering
\caption{Adroit evaluation scores and approximate performance degradation.
Each evaluation entry reports the success rate in percent, followed by
relative degradation in parentheses. Training references are estimated
from the published learning curves. Lower degradation is better.}
\label{tab:adroit_ood}
\footnotesize
\setlength{\tabcolsep}{3pt}
\renewcommand{\arraystretch}{1.15}
\begin{tabular}{lccccccc}
\hline
&
&
\multicolumn{2}{c}{\textbf{Color}} &
\multicolumn{2}{c}{\textbf{Video}} &
\multicolumn{2}{c}{\textbf{Camera View}} \\
\cline{3-4}\cline{5-6}\cline{7-8}
\textbf{Method} &
\shortstack{\textbf{Training}\\\textbf{reference}} &
\textbf{Easy} & \textbf{Hard} &
\textbf{Easy} & \textbf{Hard} &
\textbf{Easy} & \textbf{Hard} \\
\hline
SGQN  & 89 &
62 (30.34\%) & 32 (64.04\%) &
66 (25.84\%) & 32 (64.04\%) &
73 (17.98\%) & 51 (42.70\%) \\

PIE-G & 86 &
65 (24.42\%) & 60 (30.23\%) &
66 (23.26\%) & 56 (34.88\%) &
71 (17.44\%) & 16 (81.40\%) \\

SRM   & 67 &
59 (11.94\%) & 37 (44.78\%) &
50 (25.37\%) & 34 (49.25\%) &
61 (8.96\%) & 13 (80.60\%) \\

SVEA  & 51 &
42 (17.65\%) & 23 (54.90\%) &
34 (33.33\%) & 20 (60.78\%) &
42 (17.65\%) & 9.9 (80.59\%) \\

VRL3  & 90 &
0 (100.00\%) & 0 (100.00\%) &
0 (100.00\%) & 1.2 (98.67\%) &
61 (32.22\%) & 22 (75.56\%) \\

CURL  & 86 &
0.5 (99.42\%) & 0.6 (99.30\%) &
0.5 (99.42\%) & 0.3 (99.65\%) &
61 (29.07\%) & 23 (73.26\%) \\
\hline
\textbf{Shift component} & -- &
\multicolumn{2}{c}{\shortstack{
Initial-state distribution\\
$p(s_0)$}} &
\multicolumn{2}{c}{\shortstack{
Transition dynamics\\
$p(s'\mid s,a)$}} &
\multicolumn{2}{c}{\shortstack{
Observation mechanism\\
$p(o\mid s)$}} \\
\hline
\end{tabular}
\end{table*}

Table~\ref{tab:adroit_ood} shows that PIE-G's degradation increases
modestly from Easy to Hard under Color and Video, but sharply under
Camera View. SGQN shows the smallest increase under Camera View.
SRM has the smallest losses under Color and Camera View at Easy, while
PIE-G has the smallest under Video Easy ($23.26\%$ versus $25.37\%$ for
SRM). SRM's advantage weakens under Hard. Both SRM and SVEA deteriorate particularly sharply
as Camera View difficulty increases.

VRL3 and CURL exhibit near-total degradation under Color and Video at
both difficulty levels, while retaining more performance under Camera
View. Their limited variation under Color and Video therefore indicates
persistent poor performance rather than robustness.

In terms of our taxonomy, PIE-G retains performance comparatively well
as the degree of change increases within $p(s_0)$ and $p(s'|s,a)$, but
becomes substantially less robust under increasing changes in $p(o|s)$.
SGQN exhibits a complementary pattern, with greater resilience to increasing
changes in $p(o|s)$ than in $p(s_0)$ or $p(s'|s,a)$.
No single method maintains the strongest relative performance retention
across all tested components and severity levels.

% ==========================================================
\subsubsection{Robosuite}
% ==========================================================

Robosuite evaluates visual generalization and cross-embodiment transfer.
The Easy condition changes background appearance, corresponding to $p(s_0)$.
Medium additionally introduces moving illumination and changes to the arm's
color, while Hard further introduces a dynamic video background. These
two conditions therefore combine $p(s_0)$ and $p(s'|s,a)$.

Cross-embodiment transfer replaces the Panda arm with KUKA IIWA and
Kinova3. Although the OSC\_POSE controller preserves the action-space
dimension and interpretation, the changed morphology can alter the motion
produced by the same nominal action. The visible arm's appearance also
changes. Under our representation, this condition therefore combines
$p(s'|s,a)$ and $p(s_0)$.

\begin{table*}[t]
\centering
\caption{Robosuite evaluation returns and approximate performance degradation.
Each evaluation entry reports the return followed by relative degradation
in parentheses. Training references are estimated from the published
learning curves. The component row gives our structural interpretation
of each condition.}
\label{tab:robosuite_ood}
\footnotesize
\setlength{\tabcolsep}{5pt}
\renewcommand{\arraystretch}{1.15}
\begin{tabular}{lccccc}
\hline
& & \textbf{Cross Embodiment} &
\multicolumn{3}{c}{\textbf{Visual Appearance and Lighting}} \\
\cline{4-6}
\textbf{Method} &
\shortstack{\textbf{Training}\\\textbf{reference}} &
\textbf{Panda $\rightarrow$ Other Arms} &
\textbf{Easy} & \textbf{Medium} & \textbf{Hard} \\
\hline
% SGQN   & 459.76 &
SGQN   & 415.48 &
103 (75.21\%) & 257 (38.14\%) &
146 (64.86\%) & 142 (65.82\%) \\

PIE-G  & 435.68 &
103 (76.36\%) & 305 (29.99\%) &
175 (59.83\%) & 142 (67.41\%) \\

SRM    & 460.49 &
102 (77.85\%) & 275 (40.28\%) &
156 (66.12\%) & 64 (86.10\%) \\

SVEA   & 405.63 &
98 (75.84\%) & 242 (40.34\%) &
144 (64.50\%) & 75 (81.51\%) \\

DrQ-v2 & 461.44 &
76 (83.53\%) & 48 (89.60\%) &
52 (88.73\%) & 45 (90.25\%) \\

CURL   & 459.76 &
85 (81.51\%) & 59 (87.17\%) &
58 (87.38\%) & 55 (88.04\%) \\

DrQ    & 416.46 &
91 (78.15\%) & 52 (87.51\%) &
50 (87.99\%) & 50 (87.99\%) \\
\hline
\textbf{Shift component} & -- &
\shortstack{
Initial-state distribution $p(s_0)$\\
Transition dynamics $p(s'\mid s,a)$} &
\shortstack{
Initial-state distribution\\
$p(s_0)$} &
\multicolumn{2}{c}{\shortstack{
Initial-state distribution $p(s_0)$\\
Transition dynamics $p(s'\mid s,a)$}} \\
\hline
\end{tabular}
\end{table*}
Table~\ref{tab:robosuite_ood} shows that SGQN, PIE-G, SRM, and SVEA
experience larger losses when moving from Easy changes in $p(s_0)$ to
Medium conditions that additionally affect $p(s'|s,a)$. From Medium
to Hard, however, SGQN's degradation changes little and PIE-G's increases
moderately, whereas SRM and SVEA deteriorate more sharply. DrQ-v2,
CURL, and DrQ remain highly degraded across all three visual conditions. Cross-embodiment transfer involves the same combination of $p(s_0)$ and
$p(s'|s,a)$, but produces a narrower range of degradation,
approximately $75$--$84\%$ across all methods.

In terms of our taxonomy, PIE-G retains performance comparatively well
under the tested $p(s_0)$ change, while SGQN is more resilient as combined
visual changes in $p(s_0)$ and $p(s'|s,a)$ progress from Medium to Hard.
DrQ-v2, CURL, and DrQ show high degradation across all conditions.
For the other four methods, the tested conditions combining $p(s_0)$ and
$p(s'|s,a)$ consistently produce larger relative losses than the condition
affecting only $p(s_0)$. These combined changes therefore present a greater
challenge for those methods in the evaluated settings. No single method
maintains the strongest relative performance retention across all tested
conditions.

% ==========================================================
\subsubsection{CARLA}
% ==========================================================

CARLA evaluates weather and lighting changes and transfers policies trained
on highways to road layouts containing roundabouts, narrow roads, and
tunnels. Under our representation, all these conditions correspond to
$p(s_0)$. The Easy, Medium, and Hard weather settings vary the combinations
and degrees of weather and lighting changes, while scene-structure
evaluation introduces a different kind of change within the same component.

\begin{table*}[t]
\centering
\caption{CARLA evaluation returns and approximate performance degradation.
Each evaluation entry reports the return followed by relative degradation
in parentheses. Training references are estimated from the published
learning curves.}
\label{tab:carla_ood}
\footnotesize
\setlength{\tabcolsep}{6pt}
\renewcommand{\arraystretch}{1.15}
\begin{tabular}{lccccc}
\hline
& &
\multicolumn{3}{c}{\textbf{Weather and Lighting}} &
\textbf{Scene Structures} \\
\cline{3-5}
\textbf{Method} &
\shortstack{\textbf{Training}\\\textbf{reference}} &
\textbf{Easy} & \textbf{Medium} & \textbf{Hard} &
\textbf{Changed Road Layouts} \\
\hline
SGQN   & 49.18 &
36 (26.80\%) & 30 (39.00\%) &
6.4 (86.99\%) & 8.9 (81.90\%) \\

PIE-G  & 90.08 &
57 (36.72\%) & 58 (35.61\%) &
22 (75.58\%) & 12 (86.68\%) \\

SRM    & 92.43 &
90 (2.63\%) & 63 (31.84\%) &
45 (51.31\%) & 13 (85.94\%) \\

SVEA   & 90.42 &
98 ($-8.38\%$) & 65 (28.11\%) &
22 (75.67\%) & 3.3 (96.35\%) \\

DrQ-v2 & 90.41 &
38 (57.97\%) & 8.7 (90.38\%) &
3.2 (96.46\%) & $-0.1$ (100.11\%) \\

DrQ    & 38.37 &
32 (16.60\%) & 22 (42.66\%) &
23 (40.06\%) & 16 (58.30\%) \\

CURL   & 87.24 &
34 (61.03\%) & 4.7 (94.61\%) &
1.1 (98.74\%) & $-2.5$ (102.87\%) \\
\hline
\textbf{Shift component} & -- &
\multicolumn{3}{c}{\shortstack{
Initial-state distribution\\
$p(s_0)$}} &
\shortstack{
Initial-state distribution\\
$p(s_0)$} \\
\hline
\end{tabular}
\end{table*}

Table~\ref{tab:carla_ood} shows that most methods experience increasing
degradation as weather and lighting conditions progress from Easy to Hard.
SVEA and SRM retain performance well under Easy, but deteriorate as the
changes become more demanding. DrQ follows a different pattern: its loss
increases from Easy to Medium but remains similar under Hard.
DrQ-v2 and CURL already show substantial losses under Easy and approach
near-total degradation under Hard. Scene-structure changes also produce substantial losses. DrQ has the
smallest relative degradation at $58.30\%$, while all other methods
exceed $80\%$.

In terms of our taxonomy, DrQ has the lowest worst-case relative
degradation across the tested $p(s_0)$ conditions: its largest loss is
$58.30\%$, whereas every other method exceeds $85\%$ in at least one
condition. Although it does not
have the smallest loss under Easy or Medium weather and lighting, it
retains performance more consistently across the different weather
configurations and road layouts. This suggests comparatively greater
consistency across the evaluated changes within $p(s_0)$.

% ==========================================================
\subsubsection{Unitree}
% ==========================================================

The Unitree experiments evaluate policies trained under the original
visual configuration on Easy and Hard configurations involving object
color, light color, light position, and light movement. Under our
representation, these changes span static environmental configurations at reset, $p(s_0)$, and temporal environmental behavior $p(s'|s,a)$.
We therefore treat the reported conditions as combined visual changes,
rather than isolated tests of either component.

\begin{table}[t]
\centering
\caption{Unitree evaluation returns and approximate performance degradation,
aggregated over the stand and walk tasks. Each evaluation entry reports
the return followed by relative degradation in parentheses. Training
references are estimated from the published learning curves.}
\label{tab:unitree_ood}
\footnotesize
\setlength{\tabcolsep}{5pt}
\renewcommand{\arraystretch}{1.15}
\begin{tabular}{lccc}
\hline
\textbf{Method} &
\shortstack{\textbf{Training}\\\textbf{reference}} &
\textbf{Easy} & \textbf{Hard} \\
\hline
SGQN   & 625.42 & 332 (46.92\%) & 131 (79.05\%) \\
PIE-G  & 714.83 & 260 (63.63\%) & 224 (68.66\%) \\
SRM    & 690.94 & 326 (52.82\%) & 201 (70.91\%) \\
SVEA   & 735.52 & 343 (53.37\%) & 202 (72.54\%) \\
DrQ-v2 & 830.08 & 236 (71.57\%) & 89 (89.28\%) \\
CURL   & 831.5 & 253 (69.57\%) & 80 (90.38\%) \\
DrQ    & 433.56 & 204 (52.95\%) & 53 (87.78\%) \\
\hline
\textbf{Shift component} & -- &
\multicolumn{2}{c}{\shortstack{
Initial-state distribution $p(s_0)$\\
Transition dynamics $p(s'\mid s,a)$}} \\
\hline
\end{tabular}
\end{table}

Table~\ref{tab:unitree_ood} shows that all methods experience greater
degradation under Hard than Easy, but their sensitivity differs.
SGQN has the smallest loss under Easy, yet its degradation increases
sharply under Hard, second only to DrQ. PIE-G follows a more stable
pattern: its degradation changes only from $63.63\%$ to $68.66\%$,
becoming the smallest under Hard. SRM and SVEA show intermediate
increases in degradation. DrQ-v2 and CURL already have substantial
losses under Easy, which approach or exceed $90\%$ under Hard. DrQ
starts at a level similar to SRM and SVEA under Easy, but shows the
largest increase, reaching $87.78\%$ under Hard.

In terms of our taxonomy, PIE-G shows the most consistent relative
performance retention as the tested changes within the combination of
$p(s_0)$ and $p(s'|s,a)$ become more demanding. Although it does not retain
the most performance under Easy, it is less sensitive to the increase
in difficulty than the other methods. This suggests comparatively greater
resilience across the tested degrees of change within this component
combination.

% ==========================================================
\subsubsection{Robust Gymnasium}
% ==========================================================

Robust Gymnasium \cite{guROBUSTGYMNASIUMUNIFIED2025} evaluates PPO on
HalfCheetah-v4 using zero-mean Gaussian noise with standard deviations
of $0.10$ and $0.15$. Observation and action noise are applied during
evaluation, following training without noise. Reward noise is applied
during training, and the resulting policies are evaluated without noise.

% Observation noise changes $p(o|s)$, while reward noise changes
% $p(r|s,a,s')$. Action noise is added after PPO produces an action.
% Defining $a$ as the action before noise is added, this changes the
% effective transition distribution $p(s'|s,a)$.

Observation noise changes $p(o|s)$, while reward noise changes
$p(r|s,a,s')$. Action noise is added after PPO produces an action.
Since $a$ denotes the action before noise is added, action noise changes
the effective transition distribution $p(s'|s,a)$.

\begin{table*}[t]
\centering
\caption{PPO results in HalfCheetah-v4. Columns labelled ``With noise''
give the Gaussian standard deviation, with a dash indicating no noise.
Relative losses and gains are calculated against the baseline trained
and evaluated without noise.}
\label{tab:robustgym_ood}
\label{tab:robustgym_reward}
\footnotesize
\setlength{\tabcolsep}{4pt}
\renewcommand{\arraystretch}{1.15}
\begin{tabular*}{\textwidth}{@{\extracolsep{\fill}}llccccrl@{}}
\hline
& &
\multicolumn{2}{c}{\textbf{Training}} &
\multicolumn{2}{c}{\textbf{Evaluation}} &
& \\
\cline{3-4}\cline{5-6}
\textbf{Change} &
\textbf{Shift component} &
\shortstack{\textbf{Without}\\\textbf{noise}} &
\shortstack{\textbf{With}\\\textbf{noise}} &
\shortstack{\textbf{Without}\\\textbf{noise}} &
\shortstack{\textbf{With}\\\textbf{noise}} &
\textbf{Return} &
\shortstack{\textbf{Relative}\\\textbf{difference}} \\
\hline
Baseline &
-- & Yes & -- & Yes & -- & 5751 & 0.00\% \\
\hline
Observation &
Observation mechanism $p(o\mid s)$ &
Yes & -- & No & 0.10 & 901 & 84.33\% loss \\

Observation &
Observation mechanism $p(o\mid s)$ &
Yes & -- & No & 0.15 & 616 & 89.29\% loss \\
\hline
Action &
Transition dynamics $p(s'\mid s,a)$ &
Yes & -- & No & 0.10 & 3790 & 34.10\% loss \\

Action &
Transition dynamics $p(s'\mid s,a)$ &
Yes & -- & No & 0.15 & 3072 & 46.58\% loss \\
\hline
Reward &
Reward mechanism $p(r\mid s,a,s')$ &
No & 0.10 & Yes & -- & 6812 & 18.45\% gain \\

Reward &
Reward mechanism $p(r\mid s,a,s')$ &
No & 0.15 & Yes & -- & 6351 & 10.43\% gain \\
\hline
\end{tabular*}
\end{table*}
Table~\ref{tab:robustgym_ood} shows increasing degradation as the standard
deviation rises from $0.10$ to $0.15$ under both $p(o|s)$ and
$p(s'|s,a)$. Observation noise produces consistently high losses.
Action noise produces lower losses at both levels, but a larger increase
between them. This comparison applies to the tested conditions:
equal standard deviations do not necessarily represent equally strong
changes across components.

Under $p(r|s,a,s')$, both policies trained with reward noise outperform
the baseline when evaluated without noise, although the gain is smaller
at $0.15$. Higher noise therefore corresponds to lower returns between
the two tested levels in all three comparisons. However, the reward
results compare differently trained policies and do not measure
the degradation of a fixed policy exposed to reward noise during evaluation.

% ==========================================================

Across these OOD evaluations, methods show different patterns of
degradation depending on the shifted components and the specific changes
and degrees tested within them. A method may retain performance across
increasing changes in one component but deteriorate sharply in another.
Even within the same component, robustness to one tested change does not
ensure robustness to others. Robustness claims should therefore specify
both the components covered and the changes evaluated.

We next use controlled experiments to examine whether similar differences
across methods and shifted components arise under non-stationarity,
where shifts occur during ongoing interaction.

%%%%%%%%%%%%%%%%%%%%%%%%%%%%%%%%%%%%%%%%

\subsection{Distributional Shifts in the Non-Stationary Adaptation Setting: Evidence from Our Experiments}
\label{non-stationarity-empirical-result}

% Unlike the OOD setting, we did not identify existing studies that provide
% comparable component-level interventions during ongoing interaction.

Unlike the OOD setting, we did not identify existing studies that apply controlled interventions to each of the five components within the same environment during ongoing interaction. We therefore conduct controlled non-stationary experiments with DQN~\cite{mnihHumanlevelControlDeep2015},
A2C~\cite{mnihAsynchronousMethodsDeep}, and
PPO~\cite{schulmanProximalPolicyOptimization2017} in a Gridworld environment\footnote{\url{https://github.com/ardiantobow/multi-agent-rl-gridworld}}
to examine how each method responds to each component-level shift. We use
this environment because the generating function of each component can be
modified independently, so all five shift types in our taxonomy can be
implemented in the same setting. Training begins under a stationary condition
and continues after a shift is introduced at episode 250. Each experiment
modifies the generating function of one component while the others retain
their baseline specification.

% ==========================================================
\subsubsection{Environment Layout}
\label{app:env-layout}
% ==========================================================

The base environment is a deterministic $30\times30$ Gridworld with the
following properties:

\begin{itemize}
    \item \textit{Obstacles:} A set of 10 static obstacles.

    \item \textit{Goal:} A single target location.

    \item \textit{Transition model:}
    At each step, the agent selects one of five discrete actions:
    \emph{up, down, left, right}, or \emph{stay}.
    The agent moves one cell in the selected direction unless the movement
    crosses the environment boundary or collides with an obstacle, in which
    case the agent remains in its current position. The \emph{stay} action
    always preserves the current position.

    \item \textit{Observation model $p(o\mid s)$:}
    The agent receives a local field of view (FOV) of size $F\times F$
    centered on its current position. The observation contains three channels
    representing walls, obstacles, and the goal.
\end{itemize}

% ==========================================================
\subsubsection{Learning Methods}
\label{method}
% ==========================================================

We evaluate DQN~\cite{mnihHumanlevelControlDeep2015},
A2C~\cite{mnihAsynchronousMethodsDeep}, and
PPO~\cite{schulmanProximalPolicyOptimization2017}.
Each method is trained under the five component-level shifts and a
no-shift condition. Interaction and learning continue throughout both
the pre-shift and post-shift periods.

DQN learns from experience replay, while A2C and PPO use on-policy
rollouts. Within each method, the learning procedure and hyperparameters
remain consistent across conditions. This allows us to compare each
method across shift components and the three methods under each
component-level intervention.

% % ==========================================================
\subsubsection{Shift Design}
\label{subsec:nonstationary-shift-design}
% ==========================================================

We construct five controlled shifts corresponding to the component-level
sources defined in Section~\ref{subsec:component_level_taxonomy}.
Each shift changes the generating function of one selected component between
the pre- and post-shift situations. The change may affect the function's
parameters or computation. Table~\ref{tab:gridworld_shift_design} gives the pre- and
post-shift generating functions of each component in the Gridworld.

\begin{table*}[t]
\centering
\caption{Controlled component-level shifts used in the non-stationary
Gridworld experiments.}
\label{tab:gridworld_shift_design}
\footnotesize
\setlength{\tabcolsep}{6pt}
\renewcommand{\arraystretch}{1.2}

\begin{tabular}{
p{0.15\textwidth}
p{0.13\textwidth}
p{0.30\textwidth}
p{0.30\textwidth}}
\hline

\textbf{Shift Type} &
\textbf{Component} &
\textbf{Pre-shift Generating Function} &
\textbf{Post-shift Generating Function} \\

\hline

Initial-state\newline distribution
&
$p(s_0)$
&
Samples the agent's initial position on the bottom side.
&
Samples the agent's initial position on the right side, with target and
obstacle positions unchanged.
\\

\hline

Transition \newline dynamics
&
$p(s' \mid s,a)$
&
Keeps obstacles stationary.
&
Moves obstacles by one grid cell at each environment step.
\\

\hline

Reward \newline mechanism
&
$p(r \mid s,a,s')$
&
Assigns movement, obstacle, and target rewards of $(-0.0001,-10,+10)$.
&
Assigns movement, obstacle, and target rewards of $(-0.1,-1,+1)$.
\\

\hline

Observation \newline mechanism
&
$p(o \mid s)$
&
Maps the state to the local FOV without offset.
&
Maps the state to the local FOV cyclically shifted by one cell along both
spatial axes.
\\

\hline

Policy \newline implementation
&
$\pi_{\theta,\mathcal{I}}(a \mid s,o)$
&
Selects actions from unquantized Q-values (DQN) or policy logits (A2C and
PPO).
&
Selects actions from Q-values or policy logits quantized with a fixed step
size.
\\

\hline
\end{tabular}
\end{table*}

\subsubsection{Responses Across Learning Methods and Shift Components}
\label{app:implicit}

\begin{figure}[t]
\centering

\begin{subfigure}[t]{0.32\linewidth}
    \centering
    \includegraphics[width=\linewidth]{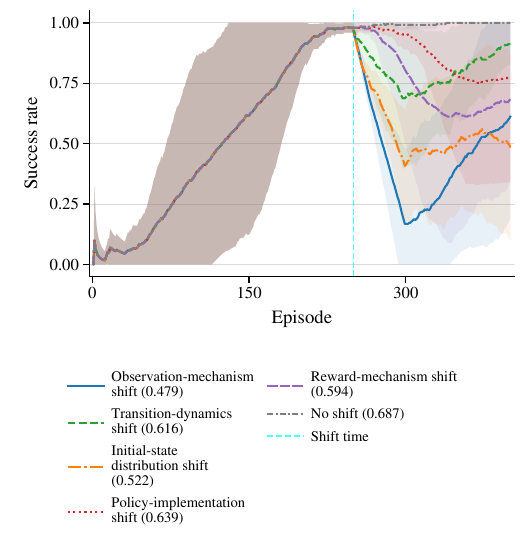}
    \caption{DQN}
    \label{fig:ns_result_dqn}
\end{subfigure}
\hfill
\begin{subfigure}[t]{0.32\linewidth}
    \centering
    \includegraphics[width=\linewidth]{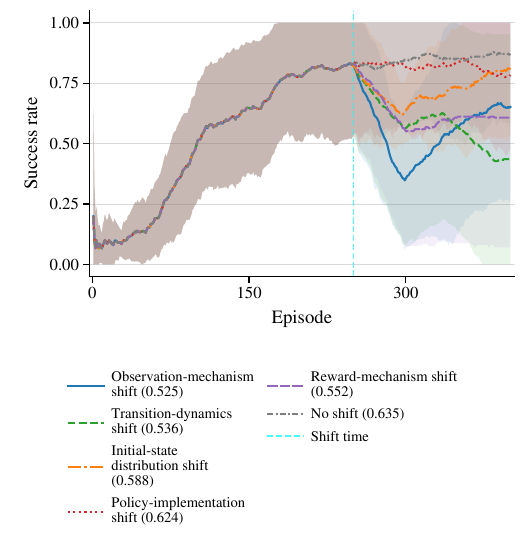}
    \caption{A2C}
    \label{fig:ns_result_a2c}
\end{subfigure}
\hfill
\begin{subfigure}[t]{0.32\linewidth}
    \centering
    \includegraphics[width=\linewidth]{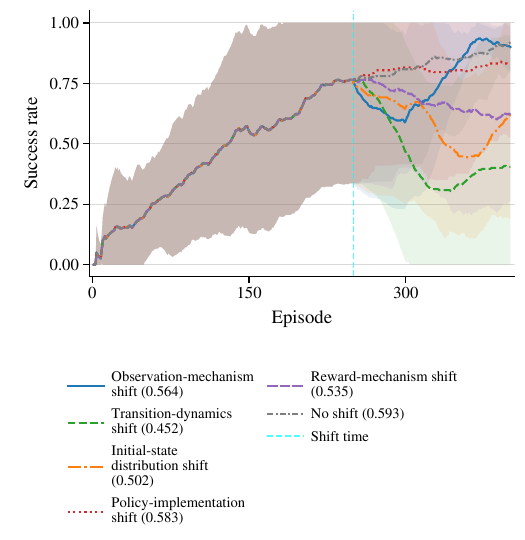}
    \caption{PPO}
    \label{fig:ns_result_ppo}
\end{subfigure}

\caption{Success-rate trajectories of DQN, A2C, and PPO under controlled
component-level shifts in the non-stationary Gridworld. The success rate of
each run is a moving average over the preceding 50 episodes.}
\label{fig:ns_result}
\end{figure}

Figure~\ref{fig:ns_result} compares DQN, A2C, and PPO under five
component-level shifts and their no-shift references, across five
independent runs. Within each method, the trajectories overlap before
episode 250 but differ after the shift. The value in parentheses after
each legend entry is the mean success rate over all 400 episodes.
Because the 250 pre-shift episodes are the same for all conditions of a
method, differences between these values arise only from the 150
post-shift episodes. For every method, all five shifted conditions fall
below the no-shift reference. Section~\ref{sec:metric-shift} examines
the post-shift response in more detail.

\begin{itemize}
    \item \textbf{DQN.} DQN recovers best from the transition-dynamics
    shift, returning close to its pre-shift performance. The
    observation-mechanism shift causes its sharpest drop, followed by
    partial recovery, and the reward-mechanism shift causes a gradual
    decline before partial recovery. After an early drop under the
    initial-state distribution shift, DQN remains substantially lower. The
    policy-implementation shift has a small immediate effect, but
    performance declines later.

    \item \textbf{A2C.} A2C remains relatively stable under the
    policy-implementation shift and, after an early drop, recovers much of
    its pre-shift performance under the initial-state distribution shift. The
    observation-mechanism shift causes a sharp drop followed by partial
    recovery, and the reward-mechanism shift causes an early drop after
    which performance remains lower. Under the transition-dynamics shift,
    A2C briefly improves before declining again and stays below its
    no-shift reference.

    \item \textbf{PPO.} PPO continues improving under the
    policy-implementation shift, although it remains below its no-shift
    reference. The observation-mechanism shift causes a smaller decline
    than in the other two methods, after which PPO returns close to its
    no-shift trajectory. The initial-state distribution shift causes a later
    decline followed by partial recovery, and the reward-mechanism shift a
    moderate, persistent decline. The transition-dynamics shift is the
    most harmful, with limited recovery.
\end{itemize}

The same shift can therefore affect the methods differently: the
transition-dynamics shift is the one DQN recovers from best but the most
harmful for PPO, and the observation-mechanism shift causes sharp drops in
DQN and A2C but only a small decline in PPO. Consistent with the OOD
examples, these results show that the response to a shift depends jointly
on the learning method and the component being changed. Robustness should
therefore be examined for each method--component combination under the
tested conditions.

%%%%%%%%%%%%%%%%%%%%%%%%%%%%%%%%%%%%%%%%%%%%%%%%%%%%%%%%%%%%%%%%%%%%%%%%
\section{Evaluation Framework for Agent Generalization or Adaptation under Distributional Shift}
\label{sec:metric-shift}
%%%%%%%%%%%%%%%%%%%%%%%%%%%%%%%%%%%%%%%%%%%%%%%%%%%%%%%%%%%%%%%%%%%%%%%%

In reinforcement learning, agent performance is typically evaluated
using aggregate measures such as cumulative return or success rate
over entire episodes or training horizons~\cite{papadopoulosExtendedBenchmarkingMultiAgent2025,papoudakisBenchmarkingMultiAgentDeep}.
While these measures characterize overall task performance, they do not
explicitly describe how an agent responds when a distributional shift
occurs. In particular, they do not separately quantify the immediate
impact of the shift, the largest subsequent degradation, the speed of
recovery, or the performance level attained after adaptation. To address this limitation, we complement the proposed taxonomy with an
evaluation framework that characterizes agent behavior around a
distributional-shift event. 

Table~\ref{tab:adaptation_framework_summary} distinguishes the framework
inputs from the computed evaluation metrics. The following subsections
define both groups in the same order as the table.

\begin{table*}[t]
\centering
\caption{Framework inputs and metrics for evaluating immediate
generalization and adaptation under distributional shifts.}
\label{tab:adaptation_framework_summary}
\footnotesize
\setlength{\tabcolsep}{6pt}
\renewcommand{\arraystretch}{1.15}
\begin{tabular}{p{0.23\textwidth}p{0.20\textwidth}p{0.49\textwidth}}
\hline
\textbf{Element} & \textbf{Symbol} & \textbf{Role} \\
\hline
\multicolumn{3}{l}{\textit{Framework inputs}} \\
\hline
Performance bounds & $M_{\mathrm{low}},\,M_{\mathrm{high}}$
& Common task-performance bounds used for normalization. \\
Window durations & $H_{\mathrm{pre}},\,H_{\mathrm{post}}$
& Evaluation durations before and after the shift. \\
Recovery target & $Z_{\mathrm{rec}}$
& Shared target on the normalized performance scale. \\
Shift time & $t_{\mathrm{shift}}$
& Last pre-shift observation, determined by the intervention
or estimated from performance changes. \\
Shift-detection threshold & $\delta_{\mathrm{detect}}$
& Minimum decrease between window means used to identify
candidate shifts in uncontrolled settings. \\
\hline
\multicolumn{3}{l}{\textit{Performance normalization and evaluation windows}} \\
\hline
Normalized performance & $Z_t$
& Performance normalized to $[0,1]$, with higher values
indicating better performance. \\
Evaluation windows & $\mathcal{W}_{\mathrm{pre}},\,\mathcal{W}_{\mathrm{post}}$
& Observation time window before and after the shift. \\
Pre- and post-shift performance & $Z_{\mathrm{pre}},\,Z_{\mathrm{post}}$
& Mean normalized performance in the respective windows. \\
\hline
\multicolumn{3}{l}{\textit{Immediate generalization}} \\
\hline
Immediate shift impact & $\Delta_{\mathrm{shift}}$
& Signed change between the last pre-shift and first post-shift
observations, before learning from post-shift experience. \\
\hline
\multicolumn{3}{l}{\textit{Adaptation performance}} \\
\hline
Mean performance change & $\Delta_{\mathrm{mean}}$
& Post-shift mean minus pre-shift mean. \\
Minimum performance & $Z_{\min}$
& Lowest normalized performance in the post-shift window. \\
Recovery time & $T_{\mathrm{rec}}$
& Time from the shift to target attainment after the minimum, set to zero if
unnecessary and to $H_{\mathrm{post}}$ if unrecovered. \\
Combined pre/post performance & $Z_{\mathrm{final}}$
& Adjusts the post-shift mean toward the pre-shift mean
using $\Delta_{\mathrm{mean}}$ and the post-shift gap to $1$. \\
End-of-window performance & $Z_{\mathrm{end}}$
& Normalized performance at the post-shift window endpoint. \\
Recovery gain & $\Delta_{\mathrm{rec}}$
& Improvement from the minimum to the endpoint:
$Z_{\mathrm{end}}-Z_{\min}$. \\
Shift adaptation score & ${\mathrm{SAS}}$
& Geometric mean $\sqrt{Z_{\mathrm{final}}Z_{\mathrm{end}}}$ divided by
$1+(T_{\mathrm{rec}}/H_{\mathrm{post}})
(1-\Delta_{\mathrm{rec}})(1-Z_{\mathrm{end}})$. \\
\hline
\end{tabular}
\end{table*}

\subsection{Framework Inputs}
\label{subsec:user-defined}

The inputs specify the performance references, evaluation periods,
recovery criterion, and location of the shift event.

\begin{itemize}
   \item \textbf{Performance bounds.}
    Select fixed task-level bounds
    $M_{\mathrm{low}}<M_{\mathrm{high}}$ that cover the evaluated
    performance range. The same bounds apply to all compared methods and
    shift events. Success rate uses bounds $0$ and $1$, whereas return bounds
    follow from the task and reward design.

    % \item \textbf{Window durations.}
    % Choose positive durations $H_{\mathrm{pre}}$ and
    % $H_{\mathrm{post}}$, measured in global training steps and shared
    % across methods. Each evaluated shift requires complete windows that
    % contain no other shift. In our experiments, these durations are
    % $10\%$ and $70\%$ of the shift interval $H_{\mathrm{phase}}$,
    % respectively.

    \item \textbf{Window durations.}
    Choose positive durations $H_{\mathrm{pre}}$ and $H_{\mathrm{post}}$
    for the windows immediately before and after each shift, measured in a
    common time unit, such as training steps or episodes, and shared across
    methods. Each evaluated shift requires complete windows that contain no
    other shift. For example, $H_{\mathrm{pre}}$ may cover the final $10\%$
    of the period before each shift, and $H_{\mathrm{post}}$ the first $70\%$
    of the period after it.

    \item \textbf{Recovery target.}
    Choose $Z_{\mathrm{rec}}\in[0,1]$ on the normalized performance
    scale defined below. The target is common to all methods on the same
    task. For example, $Z_{\mathrm{rec}}=0.90$ represents $90\%$ of
    the defined task-performance range.
    
    \item \textbf{Shift time.}
    The determination of $t_{\mathrm{shift}}$ depends on the setting:
    \begin{itemize}
        \item \textit{Controlled setting.}
        The intervention time and modified component are known from the
        experimental design.

        \item \textit{Uncontrolled setting.}
        Estimate the shift time by comparing mean performance in the
        pre- and post-shift windows defined in
        Eqs.~\ref{eq:pre_window}--\ref{eq:post_window}.
        For each candidate shift time, construct these windows and compute
        $Z_{\mathrm{pre}}$ and $Z_{\mathrm{post}}$ using
        Eqs.~\ref{eq:z_pre}--\ref{eq:z_post}. A candidate degradation
        \begin{equation}
        Z_{\mathrm{pre}}-Z_{\mathrm{post}}
        \geq\delta_{\mathrm{detect}}
        \label{eq:candidate_shift_detection}
        \end{equation}
        where $\delta_{\mathrm{detect}}\in(0,1]$ specifies the minimum
        decrease in mean normalized performance. Comparing window means
        allows detection of abrupt or gradual degradation, provided the
        decrease reaches this threshold. The detected performance decrease provides an estimate of
        $t_{\mathrm{shift}}$, but does not establish that a distributional
        shift occurred or identify which component changed.
    \end{itemize}
\end{itemize}

\subsection{Evaluation Metrics}
\label{subsec:evaluation-metrics}

We evaluate two aspects of the agent's response to a distributional shift:
\begin{itemize}
    \item \textit{Immediate generalization:} performance immediately
    after the shift, before learning from post-shift experience,
    reflecting the capability assessed in the OOD generalization setting.
    \item \textit{Adaptation:} performance and recovery as learning
    continues after the shift, reflecting the capability assessed
    in the non-stationary adaptation setting.
\end{itemize}

\begin{figure}[h]
\centering
\includegraphics[width=0.98\textwidth]{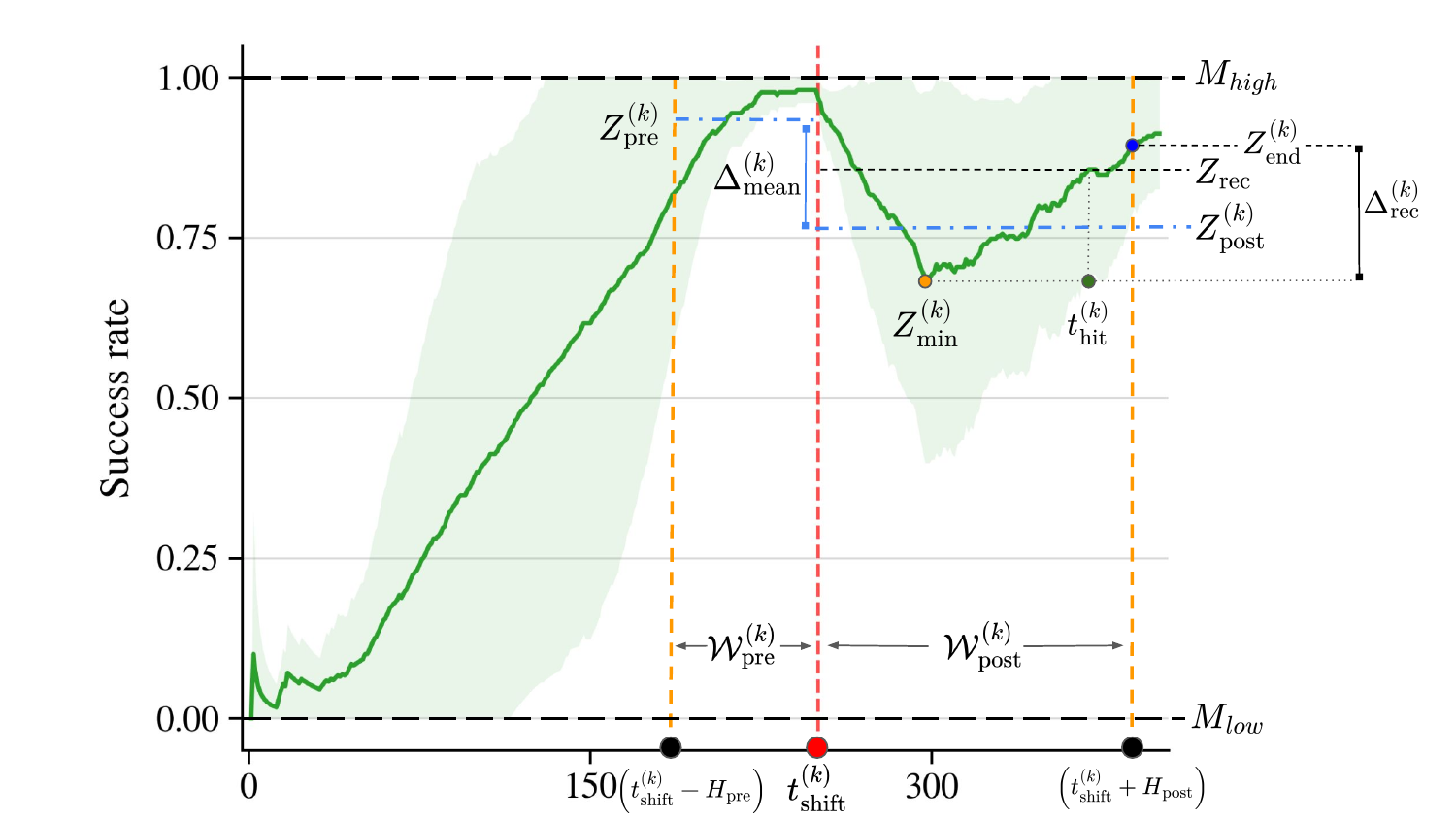}
\caption{Evaluation metric illustration.}
\label{fig:adaptation_metric_illustration}
\end{figure}

Figure~\ref{fig:adaptation_metric_illustration} illustrates the
evaluation metrics using the DQN response to a transition-dynamics
shift shown in Figure~\ref{fig:ns_result_dqn}. When several shifts occur,
the metrics are computed separately for each shift and may then be averaged
across shifts. The superscript $(k)$ in the figure denotes the $k$-th shift.
The following items define the performance scale, evaluation windows, and
individual metrics for a single shift, omitting the superscript.

\begin{itemize}
   \item \textbf{Performance normalization and evaluation windows.}
    We define the normalized performance scale and the pre- and
    post-shift windows used in the evaluation.
    \begin{itemize}
        \item \textit{Normalized performance.}
    
    Let $M_t$ denote performance measured at training step or episode $t$.
    Using the task-level bounds, define
    \begin{equation}
    Z_t=
    \begin{cases}
    \dfrac{M_t-M_{\mathrm{low}}}{M_{\mathrm{high}}-M_{\mathrm{low}}},
    & \text{if higher values are better},\\[2ex]
    \dfrac{M_{\mathrm{high}}-M_t}{M_{\mathrm{high}}-M_{\mathrm{low}}},
    & \text{if lower values are better}.
    \end{cases}
    \label{eq:absolute_performance}
    \end{equation}
    The first case applies to measures such as return or success rate, and the
    second to measures such as cost or steps to the goal. For observations within
    these bounds, $Z_t\in[0,1]$, where $0$ represents the lowest performance and
    $1$ the highest, so higher $Z_t$ always indicates better performance. This
    normalization also accommodates negative values. For example, a return of
    $-100$ within bounds $[-1000,0]$ gives $Z_t=0.90$.

        % If performance is measured by a cost, where lower values
        % are better, reverse the normalization:
        % \begin{equation}
        % Z_t=
        % \frac{M_{\mathrm{high}}-M_t}
        % {M_{\mathrm{high}}-M_{\mathrm{low}}}
        % \end{equation}
        % Thus, higher $Z_t$ always indicates better performance.

        \item \textit{Pre- and post-shift performance.}
        We place the shift between observations $t_{\mathrm{shift}}$
        and $t_{\mathrm{shift}}+1$, the last pre-shift and first
        post-shift observations, respectively. Here, $+1$ denotes
        the next training step or episode. The evaluation windows are
        \begin{equation}
        \begin{aligned}
        \mathcal{W}_{\mathrm{pre}}=\{t:\;&
        t_{\mathrm{shift}}-H_{\mathrm{pre}}<t\leq t_{\mathrm{shift}}\}
        \end{aligned}
        \label{eq:pre_window}
        \end{equation}
        \begin{equation}
        \begin{aligned}
        \mathcal{W}_{\mathrm{post}}=\{t:\;&
        t_{\mathrm{shift}}<t\leq t_{\mathrm{shift}}+H_{\mathrm{post}}\}
        \end{aligned}
        \label{eq:post_window}
        \end{equation}
        Their mean normalized performances are
        \begin{equation}
        Z_{\mathrm{pre}}
        =\operatorname{mean}(\mathcal{W}_{\mathrm{pre}})
        \label{eq:z_pre}
        \end{equation}
        \begin{equation}
        Z_{\mathrm{post}}
        =\operatorname{mean}(\mathcal{W}_{\mathrm{post}})
        \label{eq:z_post}
        \end{equation}
        Here, $\operatorname{mean}(\mathcal{W})$ denotes the mean
        of $Z_t$ over the observation times in $\mathcal{W}$.
    \end{itemize}

    \item \textbf{Immediate generalization performance.}
    This evaluation measures how performance changes immediately
    after the shift, without further learning from post-shift experience.
    \begin{itemize}
        \item \textit{Immediate shift impact.}
        Define
        \begin{equation}
        \Delta_{\mathrm{shift}}
        =Z_{t_{\mathrm{shift}}+1}-Z_{t_{\mathrm{shift}}}
        \label{eq:delta_shift}
        \end{equation}
        The first post-shift observation is collected before any
        learning update using post-shift experience. Positive values
        indicate immediate improvement, negative values indicate
        deterioration, and zero indicates preserved performance.
        This change is interpreted alongside the absolute performance
        level: preserving low performance does not imply effective
        generalization. It is reported separately from the SAS.
    \end{itemize}

    \item \textbf{Adaptation performance.}
    This evaluation characterizes performance throughout the
    post-shift period and the time required to recover.
    \begin{itemize}
        \item \textit{Mean performance change.}
        The difference between the two window means is
        \begin{equation}
        \Delta_{\mathrm{mean}}
        =Z_{\mathrm{post}}-Z_{\mathrm{pre}}
        \label{eq:delta_mean}
        \end{equation}
        Positive values indicate improvement, negative values indicate
        deterioration, and zero indicates equal means. Equal means
        do not exclude a temporary drop followed by recovery.
        This difference enters the combined performance calculation
        in Eq.~\ref{eq:z_final}.

        \item \textit{Recovery time.}
        Let $Z_{\min}$ be the lowest post-shift performance and $t_{\min}$
        the first time it occurs:
        \begin{equation}
        Z_{\min}=\min_{t\in\mathcal{W}_{\mathrm{post}}}Z_t
        \label{eq:z_min}
        \end{equation}
       
        Recovery is unnecessary when performance never falls below the
        pre-shift mean or remains at or above the recovery target $Z_{\mathrm{rec}}$.
        
        When recovery is needed, let $t_{\mathrm{hit}}$ be the first time
        at or after $t_{\min}$ that performance reaches the target:
        \begin{equation}
        t_{\mathrm{hit}}
        =\min\{t\in\mathcal{W}_{\mathrm{post}}:
        t\geq t_{\min},\ Z_t\geq Z_{\mathrm{rec}}\}
        \label{eq:t_hit}
        \end{equation}
        The recovery time is
        \begin{equation}
        T_{\mathrm{rec}}=
        \begin{cases}
        0,
        & \text{if recovery is unnecessary},\\
        t_{\mathrm{hit}}-t_{\mathrm{shift}},
        & \text{if recovery is needed and the target is reached},\\
        H_{\mathrm{post}},
        & \text{if recovery is needed and the target is not reached}.
        \end{cases}
        \label{eq:t_rec}
        \end{equation}
        Time is measured from the shift to target attainment after the
        minimum. If recovery is needed but the target is not reached within
        the post-shift window, the agent is considered unrecovered within
        that window.

        \item \textit{Combined pre/post performance.}
        This score evaluates post-shift performance while accounting for
        pre-shift performance through
        $\Delta_{\mathrm{mean}}=Z_{\mathrm{post}}-Z_{\mathrm{pre}}$:
        \begin{equation}
        Z_{\mathrm{final}}
        =
        Z_{\mathrm{post}}
        -
        \frac{1-Z_{\mathrm{post}}}
             {1+(1-Z_{\mathrm{post}})}
        \Delta_{\mathrm{mean}}
        \label{eq:z_final}
        \end{equation}
        Starting from $Z_{\mathrm{post}}$, the score moves toward
        $Z_{\mathrm{pre}}$: a lower pre-shift mean reduces the score,
        while a higher pre-shift mean raises it. Equal means retain
        their common value, distinguishing consistently high performance
        from consistently low performance.
        
        The gap $1-Z_{\mathrm{post}}$ controls this adjustment, reducing
        the influence of pre-shift performance as the post-shift mean
        approaches the normalized upper bound. The added constant $1$
        keeps the denominator positive and limits the adjustment to
        at most half the difference between the means.
        
        \item \textit{End-of-window performance.}
        The final observation in the post-shift window defines
        \begin{equation}
        Z_{\mathrm{end}}
        =
        Z_{t_{\mathrm{shift}}+H_{\mathrm{post}}}
        \label{eq:z_end}
        \end{equation}
        Unlike $Z_{\mathrm{post}}$, which summarizes the entire window,
        $Z_{\mathrm{end}}$ measures performance at its endpoint.
        
        \item \textit{Recovery gain.}
        The improvement from the lowest post-shift performance to the
        window endpoint is
        \begin{equation}
        \Delta_{\mathrm{rec}}
        =
        Z_{\mathrm{end}}-Z_{\min}
        \label{eq:delta_recovery}
        \end{equation}
        Since the endpoint belongs to the post-shift window,
        $\Delta_{\mathrm{rec}}\in[0,1]$. A larger value indicates
        a greater rebound from the minimum, with zero meaning that
        performance ends at its minimum. This gain does not itself establish
        recovery: target attainment is determined separately using
        $Z_{\mathrm{rec}}$.
        
       \item \textit{Shift adaptation score.}
        Combine pre/post performance, endpoint performance, recovery time,
        and recovery gain:
        \begin{equation}
        \mathrm{SAS}
        =
        \frac{
        \sqrt{Z_{\mathrm{final}}Z_{\mathrm{end}}}
        }{
        1+
        \displaystyle\frac{T_{\mathrm{rec}}}{H_{\mathrm{post}}}
        \left(1-\Delta_{\mathrm{rec}}\right)
        \left(1-Z_{\mathrm{end}}\right)
        }
        \label{eq:sas}
        \end{equation}
        The numerator is the geometric mean of $Z_{\mathrm{final}}$ and
        $Z_{\mathrm{end}}$, so it stays on the same scale as these two values. The
        denominator penalizes slow recovery. A larger recovery gain and a higher
        endpoint performance reduce this penalty, so slow recovery costs more when
        performance remains low at the end of the window.
            
        The score is bounded:
        \begin{equation}
        0 \leq \mathrm{SAS} \leq 1
        \label{eq:sas_bounds}
        \end{equation}
        The time penalty vanishes when recovery is unnecessary or
        $Z_{\mathrm{end}}=1$. Maintaining perfect performance throughout gives $\mathrm{SAS}=1$.

    \end{itemize}
\end{itemize}

\subsection{Application Example of the Evaluation Metric}
\label{app:adaptation_metric_usage}

\begin{figure}[h]
\centering
\begin{subfigure}[t]{0.32\linewidth}
    \centering
    \includegraphics[width=\linewidth]{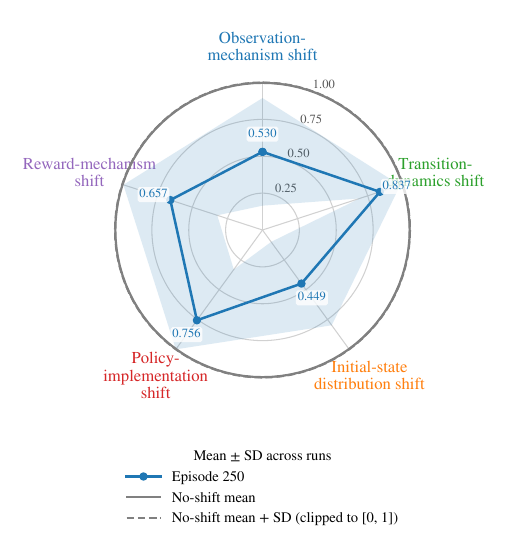}
    \caption{DQN}
    \label{fig:metric_dqn}
\end{subfigure}
\hfill
\begin{subfigure}[t]{0.32\linewidth}
    \centering
    \includegraphics[width=\linewidth]{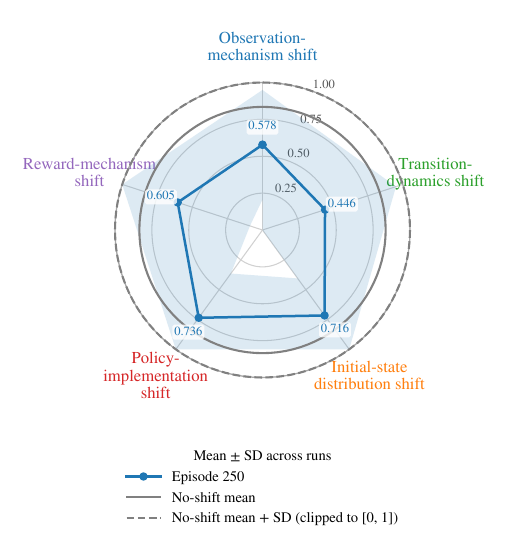}
    \caption{A2C}
    \label{fig:metric_a2c}
\end{subfigure}
\hfill
\begin{subfigure}[t]{0.32\linewidth}
    \centering
    \includegraphics[width=\linewidth]{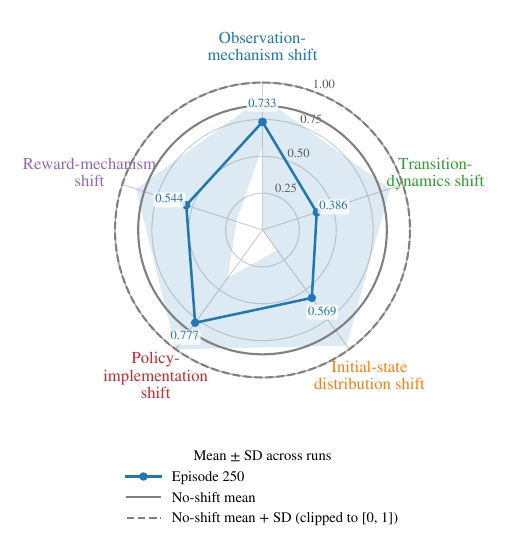}
    \caption{PPO}
    \label{fig:metric_ppo}
\end{subfigure}
% \caption{Overall adaptation scores under the five component-level
% shifts and the no-shift reference. Markers and error bars show
% the mean and standard deviation across five independent runs.}
\caption{Shift adaptation scores (SAS) under the five
component-level shifts and the no-shift reference, for the shift
introduced at episode 250.}
\label{fig:adaptation_metric_usage}

\end{figure}

We apply the proposed metric to the DQN, A2C, and PPO performance
shown in Figure~\ref{fig:ns_result} under the five component-level
shifts, using the corresponding no-shift condition as a reference.
We use success rate with bounds $M_{\mathrm{low}}=0$ and $M_{\mathrm{high}}=1$.
The shift occurs after episode 250 of 400, and we use
$H_{\mathrm{pre}}=50$ episodes ($20\%$ of the pre-shift period) and
$H_{\mathrm{post}}=150$ episodes ($100\%$ of the post-shift period), with recovery target $Z_{\mathrm{rec}}=0.90$.
Figure~\ref{fig:adaptation_metric_usage} presents the resulting shift
adaptation scores as radar charts, one per method. Each axis corresponds
to one shift condition, and the distance from the center gives the SAS on
its original $[0,1]$ scale. The blue line connects the mean scores across
five independent runs for the shift introduced at episode 250, and the
shaded region spans one standard deviation around the mean, clipped to
$[0,1]$. The solid gray reference ring indicates the mean no-shift score,
while the dashed gray ring indicates the no-shift mean plus one standard
deviation, also clipped to $[0,1]$. Greater radial distances indicate
higher SAS values, while similar distances across axes indicate more
uniform adaptation across components.
Table~\ref{tab:adaptation_metric_usage} reports the component metrics behind
each score, averaged over five independent runs.

\begin{table*}[h]
\centering
% \caption{Mean component metrics and overall adaptation scores
% for DQN, A2C, and PPO.}

\caption{Mean component metrics and shift adaptation scores (SAS) over five
independent runs for DQN, A2C, and PPO.}
\label{tab:adaptation_metric_usage}
\scriptsize
\setlength{\tabcolsep}{3pt}
\renewcommand{\arraystretch}{1.15}
\begin{tabular}{@{}lrrrrrrrrrr|c@{}}
\hline
\textbf{Shift component} &
$Z_{\mathrm{pre}}$ &
$Z_{\mathrm{post}}$ &
$\Delta_{\mathrm{shift}}$ &
$Z_{\min}$ &
$\Delta_{\mathrm{mean}}$ &
$T_{\mathrm{rec}}$ &
$Z_{\mathrm{final}}$ &
$Z_{\mathrm{end}}$ &
$\Delta_{\mathrm{rec}}$ &
& \textbf{SAS}\\
\hline
\multicolumn{12}{@{}l}{\textbf{DQN}}\\
\hline
Observation mechanism
& 0.958 & 0.440 & -0.016 & 0.168 & -0.518 & 127.6
& 0.644 & 0.612 & 0.444 && \textbf{0.530}\\
Transition dynamics
& 0.958 & 0.804 & -0.008 & 0.660 & -0.154 & 91.0
& 0.844 & 0.912 & 0.252 && \textbf{0.837}\\
Initial-state distribution
& 0.958 & 0.553 & -0.020 & 0.292 & -0.405 & 131.2
& 0.694 & 0.484 & 0.192 && \textbf{0.449}\\
Policy implementation
& 0.958 & 0.867 & 0.000 & 0.724 & -0.091 & 68.2
& 0.895 & 0.772 & 0.048 && \textbf{0.756}\\
Reward mechanism
& 0.958 & 0.747 & 0.000 & 0.456 & -0.211 & 90.0
& 0.819 & 0.684 & 0.228 && \textbf{0.657}\\

\hline
\multicolumn{12}{@{}l}{\textbf{A2C}}\\
\hline
Observation mechanism
& 0.813 & 0.556 & -0.012 & 0.332 & -0.258 & 123.2
& 0.643 & 0.652 & 0.320 && \textbf{0.578}\\
Transition dynamics
& 0.813 & 0.585 & -0.004 & 0.408 & -0.228 & 90.0
& 0.674 & 0.436 & 0.028 && \textbf{0.446}\\
Initial-state distribution
& 0.813 & 0.724 & -0.004 & 0.576 & -0.089 & 120.4
& 0.737 & 0.808 & 0.232 && \textbf{0.716}\\
Policy implementation
& 0.813 & 0.820 & 0.000 & 0.756 & 0.007 & 60.0
& 0.819 & 0.784 & 0.028 && \textbf{0.736}\\
Reward mechanism
& 0.813 & 0.626 & -0.004 & 0.532 & -0.187 & 78.4
& 0.709 & 0.608 & 0.076 && \textbf{0.605}\\

\hline
\multicolumn{12}{@{}l}{\textbf{PPO}}\\
\hline
Observation mechanism
& 0.733 & 0.756 & -0.008 & 0.540 & 0.023 & 86.6
& 0.734 & 0.900 & 0.360 && \textbf{0.733}\\
Transition dynamics
& 0.734 & 0.458 & 0.004 & 0.292 & -0.275 & 83.6
& 0.568 & 0.404 & 0.112 && \textbf{0.386}\\
Initial-state distribution
& 0.733 & 0.593 & -0.004 & 0.332 & -0.140 & 87.6
& 0.634 & 0.624 & 0.292 && \textbf{0.569}\\
Policy implementation
& 0.734 & 0.807 & 0.004 & 0.748 & 0.074 & 30.0
& 0.797 & 0.828 & 0.080 && \textbf{0.777}\\
Reward mechanism
& 0.734 & 0.679 & 0.004 & 0.500 & -0.055 & 90.0
& 0.691 & 0.620 & 0.120 && \textbf{0.544}\\

\hline
\end{tabular}
\end{table*}

\begin{itemize}
    \item \textbf{DQN.}
    The pentagon extends furthest toward transition dynamics ($0.837$)
    and policy implementation ($0.756$), and contracts toward initial-state
    distribution ($0.449$) and observation mechanism ($0.530$), where
    large mean declines combine with long recovery times.
    Policy-implementation shifts produce the higher post-shift mean,
    but their lower endpoint performance ($0.772$ versus $0.912$)
    places them below transition dynamics.

    \item \textbf{A2C.}
    The pentagon is the most even of the three, but it does not reach
    as far as DQN's at its widest point. Policy-implementation
    ($0.736$) and initial-state distribution shifts ($0.716$) score highest,
    the latter reaching endpoint performance of $0.808$ despite a long
    recovery time. Transition-dynamics shifts score lowest ($0.446$):
    endpoint performance is $0.436$, only $0.028$ above the minimum.
    Reward and observation shifts lie in between, at $0.605$ and
    $0.578$.

    \item \textbf{PPO.}
    The pentagon reaches furthest toward policy implementation
    ($0.777$), which combines an increased post-shift mean with the
    shortest mean recovery time, and toward observation mechanism
    ($0.733$), where endpoint performance reaches $0.900$. It contracts
    toward transition dynamics ($0.386$), with low post-shift and
    endpoint performance, while initial-state distribution and reward shifts
    score $0.569$ and $0.544$, respectively.
\end{itemize}

The pentagon shapes differ between methods. DQN scores highest under
transition-dynamics shifts, whereas A2C and PPO score highest under
policy-implementation shifts and lowest under transition-dynamics shifts.
Comparing A2C and PPO, A2C scores higher under initial-state distribution,
reward-mechanism, and transition-dynamics shifts, whereas PPO scores
higher under observation-mechanism and policy-implementation shifts. The
shaded regions are wide for several conditions, indicating substantial
variation across runs. Several of the differences between methods are small
relative to this variation. No single method has the highest mean score under every component: DQN leads under
transition-dynamics and reward-mechanism shifts, A2C under initial-state distribution
shifts, and PPO under observation-mechanism and policy-implementation shifts. An evaluation
restricted to one component would therefore favor whichever method suits
that component. Examining all five components identified by our taxonomy
reveals how each method responds to each shift source and supports a
fairer comparison, with robustness claims limited to the components
evaluated.

%%%%%%%%%%%%%%%%%%%%%%%%%%%%%%%%%%%%%%%%%%%%%%%%%%%%%%%%%%%%%%%%%%%%%%%%
\section{How to Apply the Taxonomy and Evaluation Framework}
\label{sec:application_protocol}
%%%%%%%%%%%%%%%%%%%%%%%%%%%%%%%%%%%%%%%%%%%%%%%%%%%%%%%%%%%%%%%%%%%%%%%%

\begin{figure}[h]
    \centering
    \includegraphics[width=\linewidth]{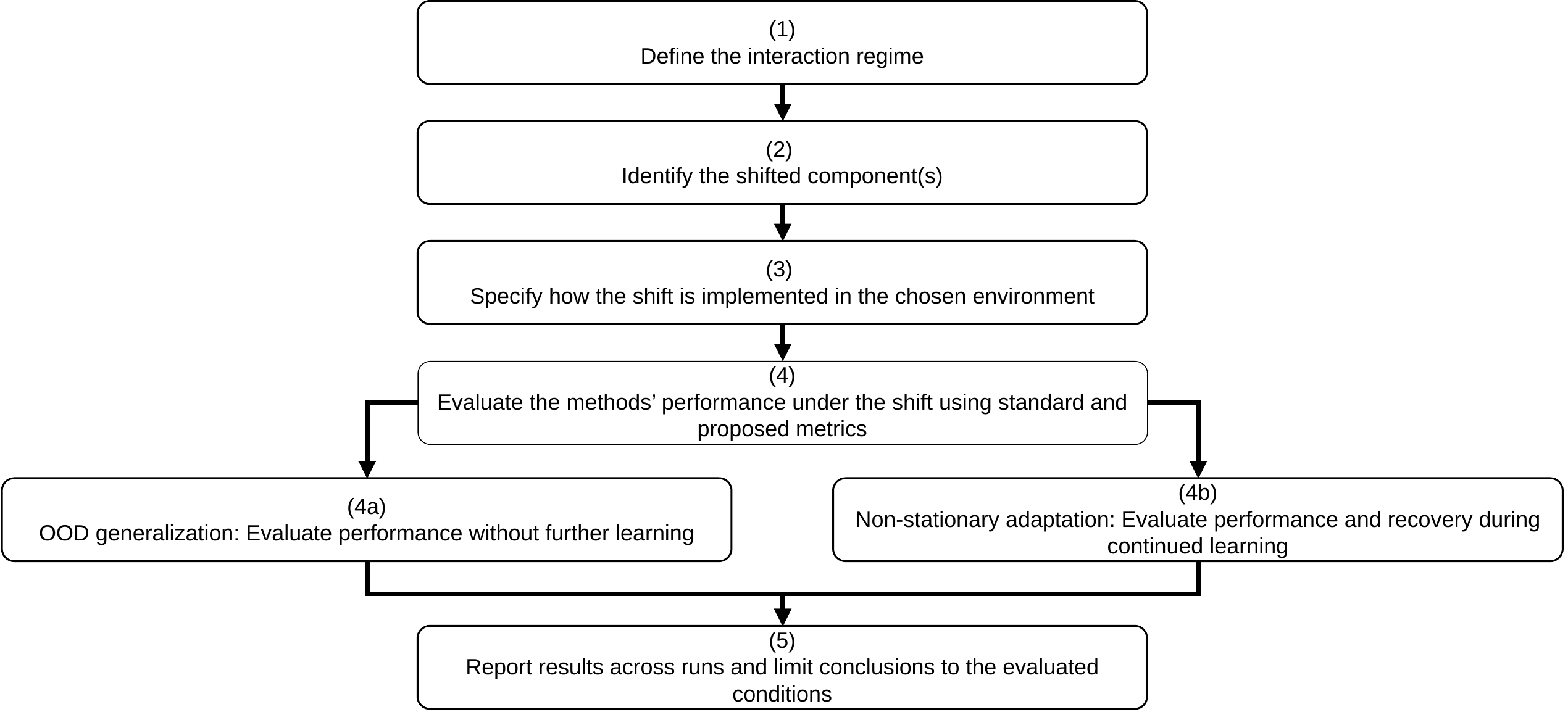}
    \caption{Steps for applying the taxonomy and evaluation framework.}
    \label{fig:application_protocol}
\end{figure}

The taxonomy guides the identification of distributional shifts,
while the evaluation framework measures the agent's response.
Figure~\ref{fig:application_protocol} outlines the steps for applying
them in a chosen environment.

\begin{enumerate}
    \item \textit{Define the interaction regime.}
    Specify when the distribution changes and whether learning
    continues afterward. In the OOD setting, the
    change occurs between training and evaluation. In the non-stationary setting, the change
    occurs during continual interaction.

    \item \textit{Identify the component(s) where the shift originates.}
    Identify whether the shift originates in the initial-state distribution,
    observation mechanism, policy implementation, transition dynamics,
    or reward mechanism, following
    Section~\ref{subsec:component_level_taxonomy}.
    If multiple components give rise to the shift, specify each one.
    
    \item \textit{Specify how the shift is implemented in the chosen environment.}
    Identify the functions, parameters, or rules that implement
    the selected component. Define their reference and shifted
    configurations, when the change takes effect, and its magnitude
    where applicable. Include a no-shift reference using the same experimental setup.

    \item \textit{Evaluate the methods' performance under the shift.}
    Select suitable standard metrics, such as return or success
    rate, and apply the proposed metrics from
    Section~\ref{subsec:evaluation-metrics}.
    The evaluation follows the selected interaction regime:
    \begin{itemize}
        \item \textit{OOD generalization.}
        Evaluate performance under shifted conditions without
        further learning. Compare absolute performance with the
        reference condition. Use $\Delta_{\mathrm{shift}}$ when
        evaluating the immediate change at the shift boundary.

        \item \textit{Non-stationary adaptation.}
        Allow learning to continue and measure performance and
        recovery over the post-shift window. 
        Report the SAS alongside its component metrics and recovery status.
    \end{itemize}

    \item \textit{Report results and delimit the conclusions.}
    Report the mean and variation across independent runs,
    comparing shifted conditions with the no-shift reference.
    Use the component metrics to explain differences in SAS values. State the interaction regimes, shifted
    components, and changes evaluated, and limit robustness claims
    to those conditions.
\end{enumerate}

%%%%%%%%%%%%%%%%%%%%%%%%%%%%%%%%%%%%%%%%%%%%%%%%%%%%%%%%%%%%%%%%%%%%%%%%
\section{Mapping Existing Methods into the Proposed Taxonomy}
\label{sec:mapping_existing_work}
%%%%%%%%%%%%%%%%%%%%%%%%%%%%%%%%%%%%%%%%%%%%%%%%%%%%%%%%%%%%%%%%%%%%%%%%

Table~\ref{tab:algo-taxonomy} maps selected studies by the distributional
shift examined, shift component level, and interaction regime.
We use these studies to illustrate the taxonomy across different shift components and interaction regimes. The selection is illustrative
rather than exhaustive and is not intended as a systematic review.

For each study, we examine the reported experimental protocol to identify
the change introduced, the component from which it originates, and whether
learning continues under the changed conditions. A change between training
and evaluation is classified as OOD, a change that occurs during ongoing
interaction while learning continues is classified as non-stationary, and a protocol with both relations is classified as combined.
Some agents adapt after an OOD change without updating the parameters
obtained during training, for example through in-context adaptation or
policy selection. Because the change occurs between training and evaluation
and these parameters are not updated under the changed conditions, the
protocols remain classified as OOD.

\begingroup
\footnotesize
\setlength{\tabcolsep}{4pt}
\renewcommand{\arraystretch}{1.15}

\begin{longtable}{@{}
>{\raggedright\arraybackslash}p{0.16\linewidth}
>{\raggedright\arraybackslash}p{0.36\linewidth}
>{\raggedright\arraybackslash}p{0.25\linewidth}
>{\raggedright\arraybackslash}p{\dimexpr0.23\linewidth-24pt\relax}@{}}

\caption{Mapping selected experimental protocols to shift components
and interaction regimes.}
\label{tab:algo-taxonomy}\\

\toprule
\textbf{Study} &
\textbf{Distributional shift examined} &
\textbf{Shift component level} &
\textbf{Interaction regime} \\
\midrule
\endfirsthead

\toprule
\textbf{Study} &
\textbf{Distributional shift examined} &
\textbf{Shift component level} &
\textbf{Interaction regime}\\
\midrule
\endhead

\midrule
\multicolumn{4}{r}{\textit{Continued on next page}}\\
\endfoot

\bottomrule
\endlastfoot

RMBPO~\cite{herremansRobustModelBasedReinforcement2024} &
Pendulum mass, torso mass, and friction coefficients are
changed during evaluation. &
Transition dynamics $p(s'\mid s,a)$ &
OOD \\
\addlinespace

CROP~\cite{altmannCROPDistributionalShiftRobust2023} &
Hole locations and maze layouts differ between training and
evaluation, while the selected observation-processing method is retained. &
Initial-state distribution $p(s_0)$ &
OOD \\
\addlinespace

Visual DR~\cite{slaouiRobustVisualDomain2020} &
Background colours differ between training and evaluation
in visual Cartpole and CarRacing. &
Initial-state distribution $p(s_0)$ &
OOD \\
\addlinespace

NPDR~\cite{muratoreNeuralPosteriorDomain2021} &
Sim--real mismatch in rotary-pendulum physical parameters, with real rollouts
updating the simulator's parameter posterior between training rounds so that
the training distribution changes over time, and with the final policies
evaluated on the real system without further learning. &
Transition dynamics $p(s'\mid s,a)$ &
Combined \\
\addlinespace

DR + adaptation~\cite{shakerimovEfficientSimtoRealTransfer2023} &
Policies trained in simulation with domain randomization are evaluated on a
real rotary pendulum carrying an extra weight that the simulator does not
model, and on simulated systems with mismatched parameters, first without
further training and then after additional training on that system. &
Transition dynamics $p(s'\mid s,a)$ &
Combined \\
\addlinespace

Context Q-learning~\cite{padakandlaReinforcementLearningAlgorithm2020} &
Transition probabilities and reward functions switch during
interaction, and context-specific policies are learned or reused. &
Transition dynamics $p(s'\mid s,a)$,
reward mechanism $p(r\mid s,a,s')$ &
Non-stationary \\
\addlinespace

BADA~\cite{liuBehaviorAwareApproachDeep2024} &
Lighting, medkit textures, and enemy counts change in ViZDoom, and the
scenario switches from defending a line in a rectangular map to defending a point in a circular map, which also changes the
ammunition supply and monster durability, while learning continues. &
Initial-state distribution $p(s_0)$, transition dynamics $p(s'\mid s,a)$ &
Non-stationary \\
\addlinespace

DiAMetR~\cite{ajayDistributionallyAdaptiveMeta2022} &
Target-location distributions (navigation, object localization, block push)
and the wind-velocity distribution (wind navigation) differ between
meta-training and meta-testing. Each meta-episode draws one task from the
fixed test distribution, with the same reward rule for every task; across
meta-episodes the agent selects a meta-policy, and within each it adapts in
context. &
Initial-state distribution $p(s_0)$,
transition dynamics $p(s'\mid s,a)$ &
OOD \\
\addlinespace

PSBL~\cite{xuMetaReinforcementLearningRobust2024} &
Test task distributions strictly shifted from training: larger navigation
domains; new target velocities in HalfCheetah; altered mass, damping,
inertia, and friction in Walker. A network with frozen parameters adapts in
context to each task drawn from the test distribution. &
Initial-state distribution $p(s_0)$,
transition dynamics $p(s'\mid s,a)$ &
OOD \\
\addlinespace

RL$^2$~\cite{duanRL$^2$FastReinforcement2016} &
Trained on $5\times5$ maze configurations and evaluated on unseen
configurations and on $9\times9$ mazes, with each trial drawing a maze and
target from the fixed test distribution and recurrent memory supporting
adaptation across episodes within a trial.&
Initial-state distribution $p(s_0)$ &
OOD \\
\addlinespace

Hybrid safe RL~\cite{hickmanHybridSafeReinforcement2025} &
After offline training on randomly collected data, 2--10 of the 15 unsafe
grid cells are relocated, and entering a hazard during online deployment
triggers a cost-model update. &
Initial-state distribution $p(s_0)$ &
Non-stationary \\

\end{longtable}
\endgroup

The mapping makes the scope of each study's robustness evaluation
explicit: which shift components and interaction regimes were tested.
It also raises a further question: would these methods remain robust
under changes originating from other components? The reported results
do not establish robustness beyond the evaluated conditions.

Within this selection, no experiment is mapped to an observation-mechanism
or policy-implementation shift. In particular, visual changes are mapped to the initial-state distribution when they modify the scene generated at reset while retaining the observation function. This highlights an opportunity to extend
evaluation to these two components, without implying that such studies
are absent from the wider literature.

%%%%%%%

%%%%%%%%%%%%%%%%%%%%%%%%%%%%%%%%%%%%%%%%%%%%%%%%%%%%%%%%%%%%%%%%%%%%%%%%
%%%%%%%%%%%%%%%%%%%%%%%%%%%%%%%%%%%%%%%%%%%%%%%%%%%%%%%%%%%%%%%%%%%%%%%%
\section{Conclusion and Discussion}
\label{sec:conclusion}
%%%%%%%%%%%%%%%%%%%%%%%%%%%%%%%%%%%%%%%%%%%%%%%%%%%%%%%%%%%%%%%%%%%%%%%%

This paper presents a unified taxonomy of distributional shifts in reinforcement
learning based on the component where a shift originates and the interaction
regime in which it occurs. It identifies five component-level sources:
initial-state distribution, observation mechanism, policy implementation,
transition dynamics, and reward mechanism. It distinguishes three interaction
regimes: OOD generalization, where the change occurs between training and
evaluation, non-stationary adaptation, where it occurs during ongoing
interaction, and a combined regime with both relations. This answers RQ1:
distributional shift is the joint mismatch caused by a component-level
change, OOD generalization and non-stationarity are the interaction regimes
in which it occurs, and discrepancies in trajectories, visitation
distributions, or returns are observed effects rather than shift sources.

For RQ2, the answer is negative. Evidence from existing benchmarks and our
controlled experiments shows that agent responses and method rankings vary
across shifted components, so strong performance under one component-level
change does not establish robustness under another, even within the same
interaction regime. Robustness claims should therefore specify the
interaction regime, components, and shift conditions evaluated.

The evaluation framework complements this classification by measuring
immediate generalization and subsequent adaptation. Its component metrics describe performance changes, recovery time, endpoint
performance, and recovery gain, while the SAS combines them into a single
score. The application protocol connects these
definitions to shift implementation and evaluation in a chosen environment.

Two directions follow from the scope of the present taxonomy. The first concerns shifts whose source is unknown. In the controlled settings considered here, the changed component is specified by design, whereas in practice only the downstream effects on interaction data and performance may be observed. Future work will establish when the component-level source can be identified from such data and when it cannot. For example, as discussed in Section~\ref{subsec:structurally_specified_shift}, a change in $p(s_0)$ may be offset by a change in $p(o \mid s)$, leaving the distribution of observations unchanged even though a distributional shift is present. Such results would move the taxonomy from specifying shifts to diagnosing them during interaction, including whether one or several components changed, and would guide the choice of adaptation strategy.

The second direction concerns offline and offline-to-online RL, where a policy learns from a fixed dataset generated by another policy. The taxonomy already identifies component-level shifts between data collection and deployment, such as changed dynamics or implementation conditions. However, the coverage gap between the dataset and the learned policy's visitation is not a component-level source. Characterizing this gap alongside component-level shifts would extend the framework to learning from fixed data.
\section*{Data and Code Availability}

The source code, experimental configuration files, and raw experimental logs required to reproduce the computational results will be made publicly available upon publication.

\section*{Acknowledgments}
Co-funded by the European Union under the Marie Skłodowska-Curie Grant Agreement No 101081465 (AUFRANDE). 
Views and opinions expressed are however, those of the author(s) only and do not necessarily reflect those of the European Union or the Research Executive Agency. 
Neither the European Union nor the Research Executive Agency can be held responsible for them. The authors would also like to acknowledge the support and collaboration of Naval Group.

\printbibliography

\end{document}